%% file: main.tex
\pdfoutput=1
\documentclass{article}

\PassOptionsToPackage{numbers,compress}{natbib}
\usepackage[main, final]{neurips_2026}

\usepackage[utf8]{inputenc} %
\usepackage[T1]{fontenc}    %
\usepackage{hyperref}       %
\usepackage{url}            %
\usepackage{booktabs}       %
\usepackage{amsfonts}       %
\usepackage{nicefrac}       %
\usepackage{microtype}      %
\usepackage{xcolor}         %

\usepackage{graphicx}
\usepackage{wrapfig}
\usepackage{algorithm}
\usepackage{algpseudocode}
\usepackage{amsmath}
\usepackage{enumitem}
\usepackage{setspace}
\usepackage{subcaption}
\usepackage{lipsum}
\usepackage{makecell}
\usepackage{pgfplots}
\pgfplotsset{compat=1.18}

\usepackage{tikz}
\definecolor{colSel}{HTML}{7F77DD}
\definecolor{colAF}{HTML}{1D9E75}
\definecolor{colOth}{HTML}{888780}

\usepackage{booktabs, multirow, xcolor, colortbl}                    \definecolor{ForestGreen}{RGB}{34,139,34}

\newcommand{\papertitle}{DisSparse}

\title{\papertitle: Pipelined Top-$p$ Sparse Attention for 
Long-Context LLM Serving}

\author{%
  Nurlan Nazaraliyev, Elaheh Sadredini, Nael Abu-Ghazaleh \\
  Department of Computer Science and Engineering \\
  University of California, Riverside \\
  \texttt{\{nnaza008, elahehs, naelag\}@ucr.edu} \\
}

\begin{document}

\maketitle
\vspace{-2em}

\begin{abstract}
Long-context LLM inference is bottlenecked by the memory demands of the key-value (KV) cache, whose size scales linearly with context length. Dynamic sparse attention reduces this bottleneck by fetching and attending only to a critical subset of tokens. The dominant primitive in dynamic sparse attention research and deployment is fixed-budget top-$k$ selection, which is favored because its deterministic block count aligns with the static memory allocation that systems like vLLM rely on. Top-$k$ uses a fixed budget regardless of how attention mass is distributed across heads, layers, and decode steps, leading to either wasted bandwidth on heads with peaky attention or dropped context on heads with diffuse attention. Top-$p$ (nucleus) selection adapts naturally to this variation but has seen limited adoption in production serving systems due to two system-level challenges: variable block counts that break deterministic block allocation and computationally expensive selection kernels that exceed the cost of attention itself. We present \papertitle{}, a serving system that integrates top-$p$ sparse attention into vLLM end-to-end. \papertitle{} decouples block selection from the scheduler's critical path, hides selection latency behind layer computation via a two-stream pipeline, and introduces a single-pass histogram-based top-$p$ kernel.  We further observe \emph{intra-block sparsity}, the inflation of selection budgets caused by mixing high- and low-importance tokens within fixed-size KV cache blocks, and address it with a token relocation kernel guided by accumulated prefill attention scores. Across long- and medium-context benchmarks on multiple models, \papertitle{} achieves 1.17$\times$ to 1.80$\times$ speedup in sparsity analysis (block selection) over state-of-the-art top-$p$ methods with negligible accuracy loss and supports up to 8$\times$ larger maximum batch sizes than dense vLLM under the same memory budget. 
Code is available at \url{https://github.com/nnurlan008/DisSparse}
\end{abstract}
\vspace{-20pt}

\input{short_intro}

\input{background}

\input{motivation}

\input{design}

\input{evaluation}

\input{conclusion}

\input{acknowledgement}

\input{appendix_new_motivation_results}

\input{appendix_A}

\input{appendix_limitations}

\input{appendix_B}

\clearpage
\newpage

\input{checklist.tex}

\end{document}

%% file: short_intro.tex
\section{Introduction}
\label{introduction}
\vspace{-10pt}

Large language models (LLMs) now support context windows spanning hundreds 
of thousands of tokens, with frontier models reaching one million or more, enabling applications such as document summarization~\cite{bai2024longbench,bai2024longbench2}, multi-turn dialogue~\cite{jeong2025accelerating}, repository-scale code generation, agentic workflows~\cite{huang2025towards}, and long-context reasoning~\cite{artificialanalysis2025lcr}. However, these advances intensify a primary bottleneck: the key-value (KV) cache that stores the conversation history must reside in fast accelerator memory, and its size grows linearly with context length. At long contexts, the KV cache dominates both memory capacity and bandwidth~\cite{sheng2023flexgen}, limiting the number of concurrent requests a serving system can sustain and the latency it can deliver.

Two complementary lines of work address this bottleneck. \emph{KV cache 
compression} reduces the per-token footprint via 
quantization~\cite{zhang2025diffkv,zandieh2026turboquant,li2025kvtuner}
or low-rank approximation~\cite{singhania2024loki,mu2025sals}. \emph{Sparse attention} reduces the number of
tokens accessed per query, either by evicting tokens from the 
cache~\cite{ge2024h2o} or by dynamically selecting 
subsets~\cite{lin2025twilight,tang2024quest}. The two axes are orthogonal: 
compression shrinks each token's representation while reading every token; 
sparsity skips tokens entirely. Combined with offloading, sparsity also 
reduces the GPU-resident footprint.

\begin{wrapfigure}[13]{r}{0.55\textwidth}
  \centering
  \includegraphics[width=\linewidth, height=2.8cm]{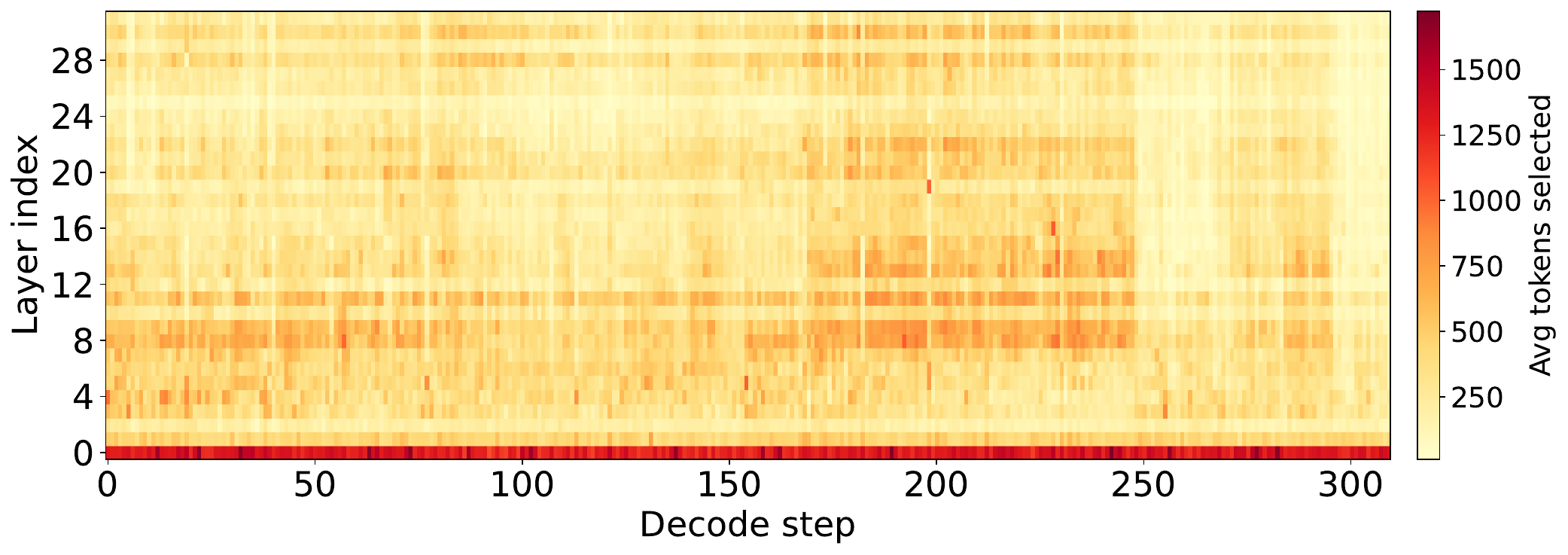}
  \caption{Llama-3.1-8B-Instruct~\cite{dubey2024llama} on NarrativeQA from LongBench~\cite{bai2024longbench}: critical token counts per layer per decode step vary substantially, showing why fixed-$k$ budgets are mismatched. Results on other models are in Appendix~\ref{label_appendix_motivation}.}
  \label{fig:attention_budget}
\end{wrapfigure}

Sparse attention divides into \emph{static} approaches that commit to a 
fixed sparsity pattern offline~\cite{zaheer2020big,beltagy2020longformer} 
and \emph{dynamic} approaches that select important tokens per query at 
runtime~\cite{chen2025magicpig,lin2025twilight,tang2024quest}. Static 
patterns are simple to implement but cannot adapt to attention distributions 
that shift across layers, heads, and decoding steps, nor to the fact that 
token importance itself evolves during generation~\cite{liu2023deja}. 
Dynamic methods address this fundamental limitation, but introduce 
system-level challenges of their own~\cite{jiang2024minference, 
levy2026dynamic}.

The dominant dynamic-sparsity primitive is \emph{top-$k$}: select a fixed number of highest-scoring blocks per head~\cite{ribar2024sparq,xiao2024infllm,sun2024shadowkv,tang2024quest,wu2025tokenselect}. Top-$k$ is attractive because $k$ is a compile-time constant; memory allocation, kernel launch parameters, and scheduling decisions can all be determined before the query arrives~\cite{tang2024quest,kwon2023efficient}. However, a fixed $k$ is mismatched to attention distributions that vary widely across layers, heads, and generation steps (Figure~\ref{fig:attention_budget}). A $k$ tuned for the densest heads wastes bandwidth on sparse heads; a $k$ tuned for the typical heads drops critical context on dense ones. Previous studies have measured these failures: fixed-budget top-$k$ violates target attention mass in over 20\% of heads~\cite{ni2026double} and degrades on tasks with diffuse attention~\cite{chen2025magicpig}.

\begin{wrapfigure}{r}{0.55\textwidth}
  \centering
  \includegraphics[width=\linewidth, height=3cm]{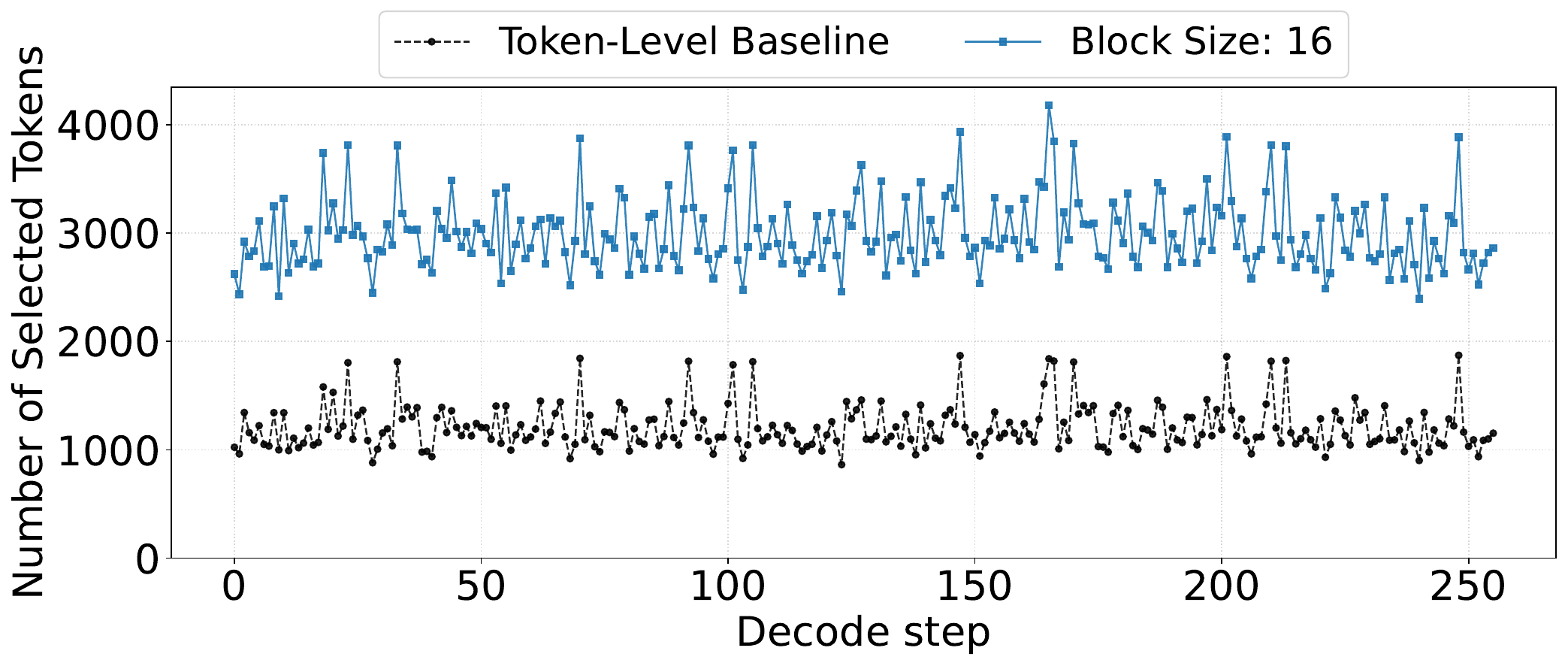}
  \vspace{-15pt}
  \caption{Llama-3.1-8B-Instruct~\cite{dubey2024llama}, Layer 0 on LongBenchV2~\cite{bai2024longbench2}. Block-level top-$p$ selects ${\sim}2\times$ more tokens than a token-level oracle to reach $p=0.95$, motivating the relocation kernel.}
  \label{fig:intra_block_sparsity}
\end{wrapfigure}

\begin{figure}[t]
  \centering
  \begin{minipage}[t]{0.48\textwidth}
  \centering
  \begin{tikzpicture}[baseline=(current axis.north)]
    \begin{axis}[
        xbar stacked,
        scale only axis,
        width=0.78\linewidth,
        height=3cm,
        bar width=2.5pt,
        xmin=0, xmax=100,
        enlarge y limits=0.10,
        xlabel={Share of layer time (\%)},
        ylabel={Batch / Context},
        xlabel style={font=\tiny, yshift=0pt},
        ylabel style={font=\tiny, xshift=0pt},
        xtick={0,25,50,75,100},
        xticklabel={\pgfmathprintnumber{\tick}\%},
        symbolic y coords={
            8/16K,8/32K,8/64K,
            16/16K,16/32K,16/64K,
            32/16K,32/32K,32/64K
        },
        ytick=data,
        y dir=reverse,
        tick label style={font=\scriptsize},
        legend style={
            at={(0.42,-0.30)}, anchor=north,
            legend columns=3, draw=none, font=\tiny,
            /tikz/every even column/.append style={column sep=5pt},
        },
        legend image code/.code={\draw[#1] (0cm,-0.06cm) rectangle (0.16cm,0.06cm);},
        tick align=outside,
        tick pos=left,
        axis y line*=left,
        axis x line*=bottom,
        major grid style={draw=gray!20},
        xmajorgrids,
    ]
    \addplot+[fill=colSel, draw=colSel] coordinates {
        (40.55,8/16K) (39.92,8/32K) (46.38,8/64K)
        (45.84,16/16K) (48.86,16/32K) (58.46,16/64K)
        (63.53,32/16K) (62.53,32/32K) (75.82,32/64K)
    };
    \addlegendentry{token selection}
    \addplot+[fill=colAF, draw=colAF] coordinates {
        (41.54,8/16K) (44.13,8/32K) (37.75,8/64K)
        (38.82,16/16K) (39.07,16/32K) (30.40,16/64K)
        (26.10,32/16K) (30.22,32/32K) (18.24,32/64K)
    };
    \addlegendentry{self-attention+FFN}
    \addplot+[fill=colOth, draw=colOth] coordinates {
        (18.02,8/16K) (16.05,8/32K) (15.87,8/64K)
        (15.43,16/16K) (12.07,16/32K) (11.14,16/64K)
        (10.37,32/16K) (7.26,32/32K) (5.94,32/64K)
    };
    \addlegendentry{other: Proj+LayerNorm}
    \end{axis}
  \end{tikzpicture}
  \caption{Per-layer time breakdown for Llama-3.1-8B-Instruct~\cite{dubey2024llama}. Selection grows from ${\sim}40\%$ to ${\sim}76\%$ of layer time as batch and sequence length scale, dominating attention and FFN combined. Measurements are based on individual kernels.}
  \label{fig:layer_breakdown}
  \end{minipage}
  \hfill
  \begin{minipage}[t]{0.48\textwidth}
  
  \centering
  \raisebox{-\height}{\includegraphics[width=\linewidth, height=3cm]{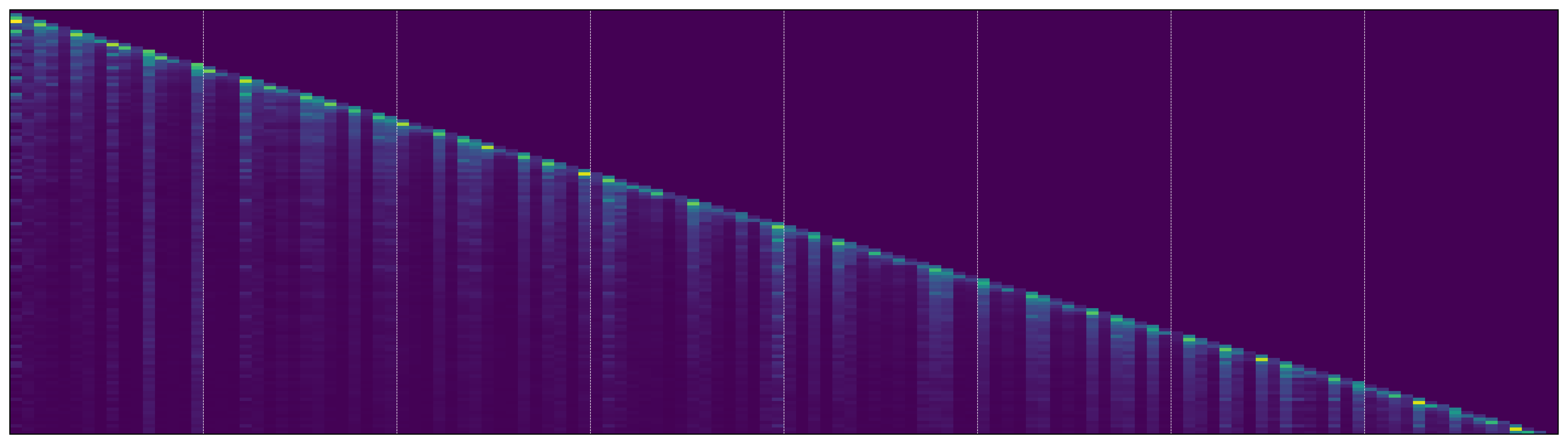}}
  \par\vspace{1.38cm}
  \caption{Layer 0 attention (Llama-3.1-8B-Instruct, LongBenchV2); white dashed lines mark 16-token blocks. Attention concentrates on a few tokens per block, yet block-level top-$p$ must load the whole block, motivating intra-block relocation.}
  \label{fig:dsa_sparsity_motivation}
  \end{minipage}
\end{figure}

\emph{Top-$p$ (nucleus) selection}, retaining exactly the blocks whose cumulative probability mass reaches threshold $p$, adapts naturally to this variation. Despite its appeal, top-$p$ has seen little adoption in production serving systems due to two challenges: \textbf{(1)} existing top-$p$ implementations are computationally expensive, locating the exact threshold via multi-pass search~\cite{lin2025twilight} or accumulating partial attention until $p$ is reached~\cite{zhou2025progressive}; and \textbf{(2)} the number of selected blocks is unknown until selection completes, complicating the deterministic memory allocation that schedulers like vLLM rely on. Both increase selection cost, as Figure~\ref{fig:layer_breakdown} shows, top-$p$ selection~\cite{lin2025twilight} starts dominating layer execution time at batch sizes $\geq 16$.

In this paper, we present \papertitle{}, a system that integrates top-$p$ sparse attention into vLLM~\cite{kwon2023efficient} end-to-end. \papertitle{} addresses these challenges via co-designed components. \textit{For challenge (1)}, we introduce a histogram-based kernel that computes an approximate threshold in a single pass over $N$ tokens using $B$ histogram bins, replacing multi-pass binary search. \textit{For challenge (2)}, our key insight is that top-$p$ selection can be \emph{decoupled from the scheduler's critical path}: \papertitle{} delegates retrieval of already-computed KV blocks to the model executor, leaving only new-block allocation in the scheduler. To overlap the resulting selection latency with computation, we organize each decoding step into two CUDA streams: a \emph{selection stream} that runs top-$p$ analysis on an INT4 KV cache replica~\cite{lin2025twilight} for layer~$\ell{+}1$ and prefetches the selected blocks, and a \emph{compute stream} that executes self-attention and the FFN for layer~$\ell$, building on high cosine similarity of consecutive layers' hidden states~\cite{lee2024infinigen}.

Beyond these two challenges, we observe an additional inefficiency specific to block-paged KV caches: tokens within a block often have highly skewed importance, so block-level top-$p$ must select more blocks than a token-level oracle to reach the same $p$ (Figure~\ref{fig:intra_block_sparsity}). We call this \textit{intra-block sparsity} and introduce a relocation algorithm that re-clusters tokens post-prefill by their accumulated prefill attention scores. While no single relocation can perfectly predict importance at every future decode step, tokens with high cumulative prefill scores tend to remain important during generation~\cite{liu2023scissorhands,li2024snapkv}; grouping them into shared blocks reduces the number of selected blocks by up to 14\%.

We evaluate \papertitle{} on long-context~\cite{bai2024longbench,bai2024longbench2} and medium-context~\cite{cobbe2021training,reddy2019coqa,rae2019compressive} benchmarks across Llama~3.1-8B-Instruct~\cite{dubey2024llama}, LLaMA2-7B-Chat~\cite{touvron2023llama}, Longchat-7B-v1.5-32k~\cite{li2023longchat}, Qwen2.5-7B-Instruct~\cite{yang2025qwen25}, and Phi-3-mini-128k-instruct~\cite{abdin2024phi3}. \papertitle{} achieves 1.17$\times$ to 1.80$\times$ speedup over Twilight~\cite{lin2025twilight}, the state-of-the-art top-$p$ method, with negligible accuracy loss. At the system level, \papertitle{}'s integration into vLLM~\cite{kwon2023efficient} increases the maximum sustainable batch size by up to 8$\times$ over dense vLLM under the same GPU memory budget.

\noindent Our contributions are:
\begin{enumerate}[leftmargin=*, itemsep=2pt, topsep=2pt]
    \item A \textbf{pipelined top-$p$ sparse attention system} that decouples block selection from the scheduler's critical path and overlaps selection and prefetch with layer computation via a two-stream design, integrating into vLLM~\cite{kwon2023efficient} with minimal scheduler modifications (\S\ref{system_1}).

    \item A \textbf{histogram-based approximate top-$p$ kernel} that replaces multi-pass binary search~\cite{lin2025twilight} or progressive accumulation~\cite{zhou2025progressive} with a single $O(N)$ pass using $O(B)$ registers (\S\ref{para:fused_kernel}).

    \item A \textbf{token relocation kernel} that re-clusters KV pairs by accumulated prefill attention scores, reducing the number of blocks selected by top-$p$ by up to 14\% (\S\ref{para:relocation}).

    \item A \textbf{comprehensive evaluation} on long- and medium-context benchmarks across multiple models, demonstrating up to 1.80$\times$ selection speedup over Twilight~\cite{lin2025twilight} with negligible accuracy degradation and up to 8$\times$ batch-size increase over dense vLLM (\S\ref{eval_section}).
\end{enumerate}

%% file: background.tex
\section{Background and Related Work}
\label{related_work}

\textbf{Sparse Attention.}
Attention distributions in LLMs are inherently skewed: a small fraction of tokens dominates the attention mass~\cite{zhang2023h2o,liu2023deja}, and the majority can be skipped during decoding with negligible accuracy loss~\cite{xiao2024streamingllm}.
\textit{Static} approaches (e.g., sliding window, dilated attention) apply fixed token-dropping patterns but typically require training or fine-tuning with the fixed pattern~\cite{beltagy2020longformer}.
\textit{Dynamic} sparse attention (DSA) instead adapts on the fly to input context. Within DSA, \textit{training-aware} methods modify the architecture and require continued pre-training~\cite{liu2025deepseek}, while \textit{training-free} methods~\cite{tang2024quest,chen2024arkvale,xiao2024infllm} operate at inference time on off-the-shelf models. The dominant primitive in training-free DSA is \textit{top-$k$}~\cite{tang2024quest,xiao2024infllm,sun2024shadowkv,wu2025tokenselect}, but it cannot adapt to the wide variation in attention distributions across layers, heads, and decoding steps, leading to accuracy degradation on tasks with diffuse attention patterns~\cite{chen2025magicpig,ni2026double}.

\textbf{Top-$p$ Sparse Attention.}
\textit{Top-$p$ (nucleus)} selection retains the minimal set of tokens whose cumulative attention mass exceeds threshold~$p$. Twilight~\cite{lin2025twilight} first selects candidate pages via a fixed top-$k$ budget, then applies token-level top-$p$ pruning through binary search; the multi-pass search incurs significant selection overhead and requires multiple separately launched kernels (GEMV, softmax, and top-$p$ search). Progressive Sparse Attention~\cite{zhou2025progressive} processes blocks in ranked order and accumulates mass until the threshold is met, but its iterative microbatch execution incurs substantial CPU--GPU synchronization overhead. Double-P~\cite{ni2026double}, concurrent work, performs hierarchical top-$p$ estimation via k-means centroids followed by token-level refinement. These methods advance the algorithmic design of top-$p$ but do not address the systems-level challenges of integrating variable-count selection into production serving frameworks; \papertitle{} fills this gap by co-designing top-$p$ with the scheduler, memory manager, and execution pipeline.

\textbf{KV Cache Compression.}
Quantization reduces the per-token KV cache footprint, with fixed-precision~\cite{lin2025qserve,zhao2024atom} and importance-adaptive~\cite{zhang2025diffkv} variants. A related but distinct line, eviction, permanently removes unimportant tokens: H2O~\cite{zhang2023h2o} and Scissorhands~\cite{liu2023scissorhands} drop tokens with low accumulated attention, StreamingLLM~\cite{xiao2024streamingllm} retains only initial tokens plus a sliding window, and SnapKV~\cite{li2024snapkv} compresses during prefill. While effective at reducing capacity, permanent eviction can destroy the context needed for long-range retrieval. Both are orthogonal to \papertitle{} and can be composed with our sparse attention to yield cumulative gains.

\textbf{Serving Systems and Asynchronous KV Management.}
Modern serving frameworks~\cite{kwon2023efficient,zheng2024sglang,zhong2024distserve,shi2025nexus} manage the KV cache at block granularity via PagedAttention~\cite{kwon2023efficient}, which assumes the per-request block count is known at scheduling time, an assumption top-$p$ violates. These systems support either \textit{hybrid} (colocated) or \textit{disaggregated}~\cite{zhong2024distserve,shi2025nexus} prefill/decode; \papertitle{} adopts disaggregation to isolate decode-stage improvements but is compatible with both. To overlap data movement with computation, InfiniGen~\cite{lee2024infinigen} prefetches KV blocks speculatively by exploiting the high cosine similarity between hidden states at consecutive layers, an observation consistent with broader evidence of consecutive layer representations' similarity in LLMs~\cite{men2025shortgpt}. \papertitle{} builds on this insight: a two-stream pipeline runs top-$p$ analysis for layer~$\ell{+}1$ alongside layer~$\ell$'s attention and FFN, hiding selection and prefetch latency behind useful computation.

%% file: motivation.tex
\section{The Systems Challenge of Dynamic Top-$p$ Sparsity}
\label{problem}

\begin{wrapfigure}{r}{0.52\textwidth}
    \centering
    \includegraphics[width=0.52\textwidth]{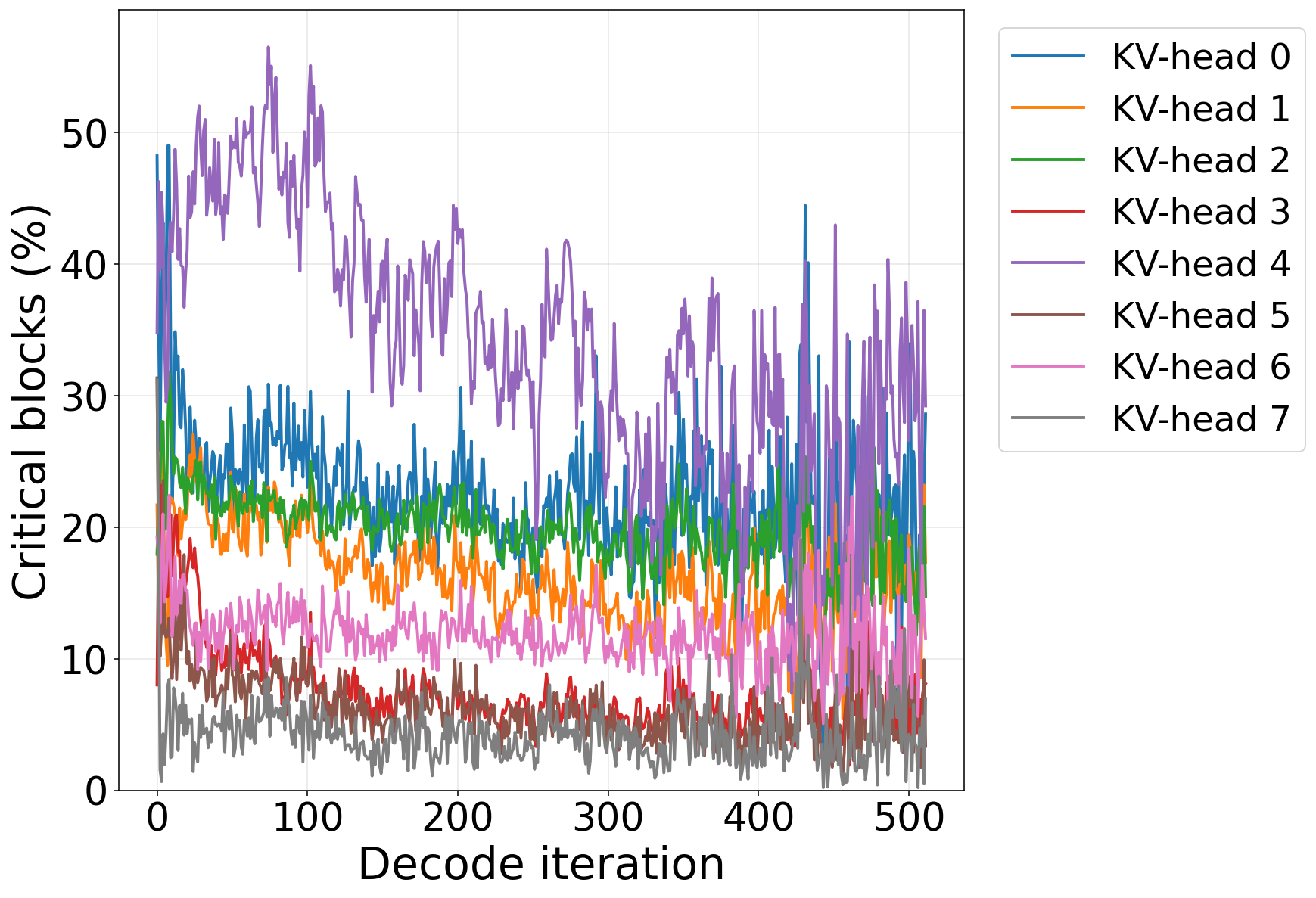}
    \caption{Percentage of selected critical blocks per KV head for layer 4 in Llama-3.1-8b-Instruct on LongBenchV2 samples. Results on other models are in Appendix~\ref{label_appendix_motivation}.}
    \label{fig:dsa_sparsity_motivation_head}
\end{wrapfigure}

While dynamic sparse attention (DSA) offers a promising path to reducing KV cache memory bandwidth during autoregressive decoding, integrating it into production serving frameworks introduces substantial systems-level friction~\cite{levy2026dynamic}. To date, system-level efforts have largely converged on top-$k$ selection~\cite{tang2024quest,sun2024shadowkv} because its fixed budget aligns with the deterministic memory allocation in modern schedulers.

However, top-$k$'s fixed selection count is mismatched to the attention dynamics of modern LLMs. Figure~\ref{fig:dsa_sparsity_motivation_head} shows that attention distributions vary substantially across KV heads within a decode step and across steps for the same head. Prior work attributes this to head-level specialization: heads are either \textit{retrieval} (gathering information from the full context) or \textit{local} (focusing on nearby tokens)~\cite{wu2024retrieval,xiao2024duoattention}. A static $k$ over-selects on focused heads, wasting bandwidth on irrelevant blocks, and under-selects on dispersed ones, discarding blocks that contribute to output quality.

Top-$p$ (nucleus) selection can address this mismatch by adapting the budget per layer and head~\cite{desai2025vattention,lin2025twilight}, but faces the following challenges:

\textbf{Challenge 1: Cost of Relative Score Computation.} Top-$k$ can rank blocks from compact metadata such as element-wise bounds~\cite{tang2024quest} or query-conditioned approximations~\cite{ribar2024sparq}, never touching the full cache. Top-$p$, by contrast, must compute and accumulate the actual attention scores across all cached tokens to locate where the cumulative threshold is met. Performing this on the full-precision (FP16) cache would require loading the data that sparse attention is designed to skip, negating the bandwidth reduction entirely. A practical alternative is to maintain a lightweight, quantized (INT4) replica of the KV cache dedicated to score computation. However, this introduces additional memory overhead and requires careful pipelining to avoid serializing the score computation with the main attention kernel (\S\ref{system_1}).

\textbf{Challenge 2: Non-Deterministic Memory Demand.} State-of-the-art frameworks such as vLLM rely on deterministic resource provisioning: the scheduler pre-allocates KV cache blocks from a shared free pool before execution, assuming the per-request block count is known at scheduling time. Top-$p$ violates this assumption: the number of selected blocks is determined only after computing attention scores and varies sharply across layers and heads within a single decode step (Figure~\ref{fig:attention_budget}). The scheduler must then either over-provision conservatively, wasting memory and shrinking batch sizes, or defer allocation to the model runner at runtime, breaking the separation between scheduling and execution. Neither is acceptable without a principled change to the allocation interface.

Beyond these deployment challenges, block-level granularity introduces an additional inefficiency. Modern inference engines manage the KV cache in blocks of 16 (32, 64, etc.) tokens via PagedAttention~\cite{kwon2023efficient} to avoid memory fragmentation. Because attention scores within a block are often highly skewed, a single critical token forces retrieval of its entire enclosing block (Figure~\ref{fig:dsa_sparsity_motivation}), inflating the block count needed to reach threshold $p$ well beyond the count of individually critical tokens. Figure~\ref{fig:intra_block_sparsity} quantifies this gap: block-level top-$p$ requires $\sim2\times$ more tokens than a token-level oracle for layer 0 of Llama-3.1-8b-Instruct on a LongBenchV2~\cite{bai2024longbench2} sample. The effect is worst under focused attention distributions, precisely the regime where top-$p$ should yield the greatest savings.

These two challenges are interconnected: the variable block count from top-$p$ (Challenge 2) compounds the cost of the score computation that produces it (Challenge 1), and intra-block sparsity further amplifies the bandwidth pressure both create. Addressing them in isolation yields partial solutions; effective deployment requires co-designing the selection algorithm with the memory management substrate.

\begin{figure}[t]
  \centering
  \begin{minipage}[t]{0.36\textwidth}
    \vspace{0pt}
    \centering
    \includegraphics[width=\linewidth]{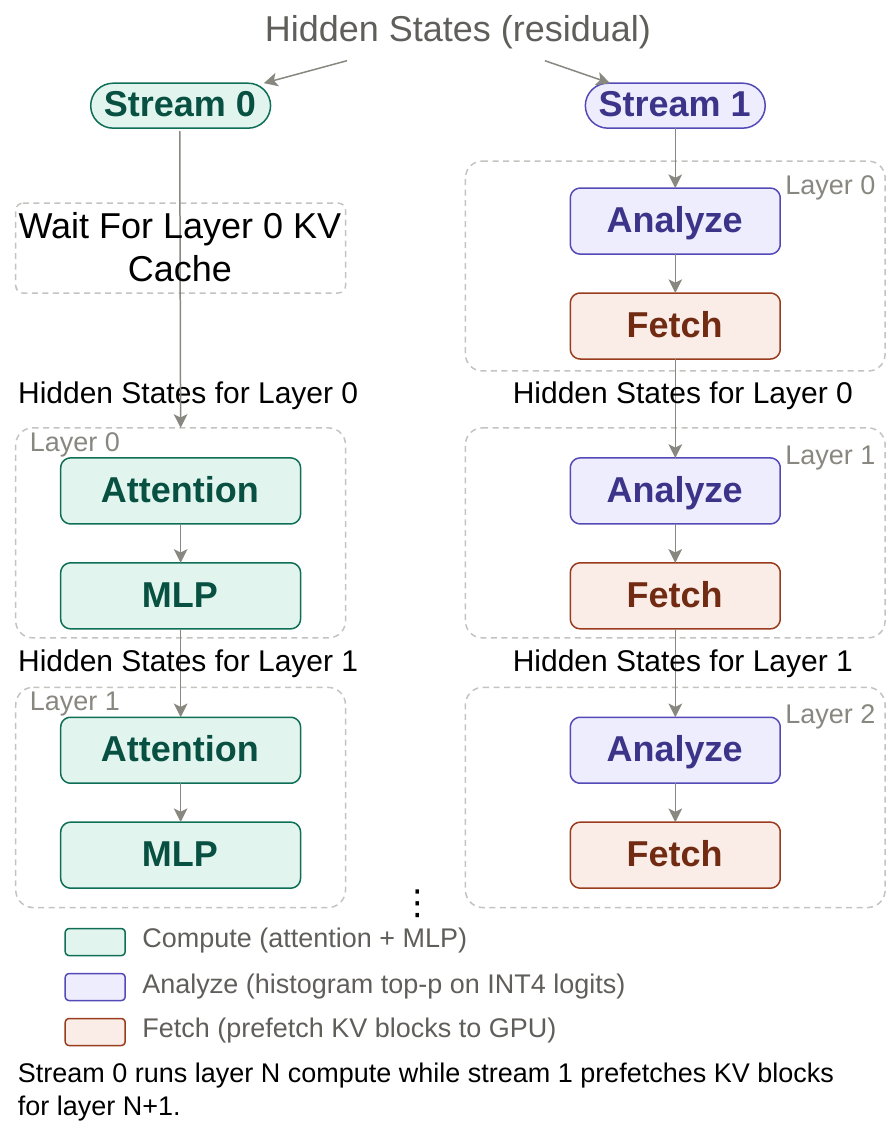}
    \caption{Pipelined Execution}
    \label{fig:kv_cache_pipeline}
  \end{minipage}
  \hfill
  \begin{minipage}[t]{0.61\textwidth}
    \vspace{0pt}
    \input{algorithms/top_p_gemini}
  \end{minipage}
\end{figure}

%% file: algorithms/top_p_gemini.tex
\begin{minipage}{\linewidth}
\hrule height 0.8pt
\vspace{4pt}
\refstepcounter{algorithm}
\noindent\textbf{Algorithm \thealgorithm}. Paged INT4 Scoring + Top-$p$
\label{alg:fused_attn_short}
\vspace{4pt}
\hrule height 0.4pt
\vspace{4pt}
\scriptsize
\linespread{0.9}\selectfont
\begin{algorithmic}[1]
\Procedure{TopPSelect}{$Q, K_{\text{INT4}}, P_{\text{tgt}}, k, B$}
    \Statex \textbf{Phase 1 (scoring kernel): Per-block LSE} (context split across thread blocks)
    \State $m \gets -\infty$,\; $l \gets 0$
    \For{each tile $j$ of $K_{INT4}$-cache}
        \State $S_j \gets (Q K_j^\top) \cdot s$ \Comment{INT4 dequant, optional ALIBI/softcap}
        \State $V_j \gets \max(S_j) + \log_2\!\sum \exp_2(S_j - \max(S_j))$ \label{alg:line:block_lse}
        \State Update streaming $(m, l)$;\; store $V_j$ to block-LSE buffer
    \EndFor

    \Statex \textbf{Phase 2 (top-$p$ kernel): Top-$k$, then threshold $\tau$ via fast-path or histogram}
    \State $L \gets m + \log_2 l$ \Comment{global LSE}
    \State $T \gets \mathrm{TopK}(\{V_j\}_j,\, k)$ \Comment{$k$ rounds of block-parallel argmax}
    \State $P_{\text{top}} \gets \sum \exp_2(T - L)$,\; $\hat{V}_{\min} \gets \min(T) - L$
    \If{$P_{\text{top}} \ge P_{\text{tgt}}$}
        \State $\tau \gets \hat{V}_{\min} - \epsilon$ \Comment{Fast path} \label{alg:line:fast_path}
    \Else \Comment{Histogram path}
        \State $P_{\text{above}} \gets \sum_{j:\, V_j - L > \hat{V}_{\min}} \exp_2(V_j - L)$
        \State Build size-$B$ histogram $H$ of $\exp_2(V_j - L)$ for $V_j - L \in [\hat{V}_{\min} - W,\, \hat{V}_{\min}]$
        \State $b \gets \max\{i : \mathrm{RevCumSum}(H)_i \ge P_{\text{tgt}} - P_{\text{above}}\}$
        \State $\tau \gets (\hat{V}_{\min} - W) + (b/B) \cdot W$ \Comment{Slow path}
    \EndIf
    \State \Return blocks $j$ with $V_j - L \ge \tau$ (plus sink and local-window blocks)
\EndProcedure
\end{algorithmic}
\vspace{4pt}
\hrule height 0.8pt
\end{minipage}

%% file: design.tex
\section{\papertitle}
\label{solution}

\subsection{System Architecture and Design}
\label{system_1}

\papertitle{} uses a disaggregated serving architecture~\cite{vllm_disagg_prefill,zhong2024distserve}, executing prefill and decode on separate devices to isolate decode-stage optimizations from prefill interference. We observe that, on long-context inputs (16K average sequence length), decode accounts for $\sim$78\% of total inference time on a 2-GPU disaggregated vLLM setup, and this fraction grows with sequence length.

\papertitle{} maintains a \textit{metadata cache}, an INT4 replica of the main FP16 key cache (8$\times$ smaller), used exclusively by the sparse analyzer for top-$p$ block selection. After the prefill engine generates the first token and its KV cache, the metadata cache is transferred to the decode engine via asynchronous NCCL collectives~\cite{nvidia_nccl}. During each decode step, the executor concurrently fetches selected FP16 KV blocks on the transfer stream while running the forward pass on the compute stream.

\begin{wrapfigure}[15]{r}{0.35\columnwidth}
    \centering
    \vspace{-6.0mm}
    \includegraphics[width=0.33\columnwidth]{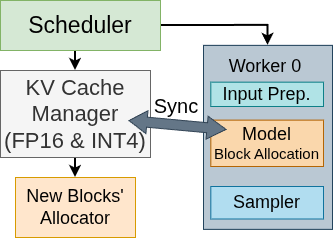}
    
    \caption{vLLM System Overview. \papertitle{} adds the new INT4 cache to the KV cache manager and synchronization unit between the workers (model executor) and the KV cache manager (scheduler).}
    \label{fig:dsa_memory_allocator}
\end{wrapfigure}

\paragraph{Hybrid block allocation.} 
Standard vLLM uses a single scheduler-owned block allocator for newly generated tokens. Sparse attention with offloading introduces a second pattern: allocations for already-computed remote blocks fetched per step by top-$p$. The scheduler cannot model these efficiently as their count and identity are determined at execution time, not scheduling time. \papertitle{} therefore splits allocation between the scheduler and the device-side executor (Figure~\ref{fig:dsa_memory_allocator}): the scheduler allocates blocks for newly generated tokens and the metadata cache, while the executor allocates blocks for transient sparse fetches. The two allocators synchronize at each step's end; the scheduler retains lifecycle responsibility (memory release, prefix caching). When the executor's free pool is insufficient, affected requests stall via a per-batch boolean mask without changing the batch configuration.

\paragraph{Computation-communication overlap.} 
Computing top-$p$ on the INT4 replica solves Challenge~1's bandwidth 
problem but introduces a new requirement: selection for layer~$\ell$ 
must complete before its attention can fetch the chosen blocks, or the 
forward pass stalls. \papertitle{} avoids this by exploiting prior 
observations~\cite{lee2024infinigen,men2025shortgpt} that consecutive 
layers' hidden states are highly similar; across our evaluated models, 
the cosine similarity ranges from 0.87 to 0.95\footnote{0.878 ($\pm$0.002) for Llama-2-7B-Chat~\cite{touvron2023llama} at 4K, 0.891 ($\pm$0.001) for Longchat-7B-v1.5-32k~\cite{li2023longchat}, 0.868 for Llama-3.1-8B-Instruct~\cite{dubey2024llama}, 0.930 ($\pm$0.001) for Phi-3-mini-128k-instruct~\cite{abdin2024phi3}, and 0.951 ($\pm$0.001) for Qwen3-32B~\cite{yang2025qwen3} at 32K.}, so the 
previous layer's hidden state reliably predicts the next layer's 
selection. Our two-stream pipeline (Figure~\ref{fig:kv_cache_pipeline}) 
runs one layer ahead of compute. The \emph{selection stream} runs 
top-$p$, fetches remote blocks, and allocates local slots for 
layer~$\ell{+}1$ using a query computed from layer~$\ell$'s hidden 
state through layer~$\ell{+}1$'s $W_Q$. The \emph{compute stream} runs 
layer~$\ell$'s forward pass: self-attention over its previously 
selected blocks plus the surrounding projections, normalizations, FFN, etc.

\subsection{Kernels Implementation}
\label{system_2}

~

\input{algorithms/token_relocation_gemini_version}

\paragraph{INT4 Scoring and Top-$p$ Kernels.}
\label{para:fused_kernel}
\papertitle{} addresses Challenge~1 with two separate kernels: an INT4 GEMV
kernel that scores blocks directly on the INT4 cache, and a histogram top-$p$
kernel that selects blocks from those scores in a single pass, replacing
the multi-pass binary search of prior top-$p$ implementations. Both kernels
run on the selection stream (\S\ref{system_1}); a condensed version is
shown in Algorithm~\ref{alg:fused_attn_short} (full version in
Appendix~\ref{label_appendix_B}). The scoring kernel extends the tiling and
online-softmax patterns of FlashAttention-2~\cite{dao2024flashattention2},
with grid axes indexed by query and KV head, and the top-$p$ kernel
identifies critical KV cache blocks via a histogram approximation of top-$p$
over per-block log-sum-exp (LSE) scores.

In Phase~1, the scoring kernel streams over $K_{\text{INT4}}$-tiles, computes 
raw attention logits $QK_{\text{INT4}}^\top$, and aggregates them into 
per-block LSE values (line~\ref{alg:line:block_lse}), avoiding explicit 
per-token softmax. In parallel, it maintains a streaming buffer of the 
$k$ largest block LSEs (we use $k=8$). In Phase~2, the top-$p$ kernel checks 
whether these Top-$k$ blocks already cover the target mass 
$P_{\text{tgt}}$. If so, the fast path sets $\tau$ just below 
$\hat{V}_{\min}$ (line~\ref{alg:line:fast_path}), selecting the Top-$k$ 
blocks. Otherwise, the slow path constructs a $B=64$-bin histogram of 
block probabilities over the log-probability window 
$[\hat{V}_{\min} - W,\, \hat{V}_{\min}]$ ($W = 8$, covering a 
$2^8\times$ probability range below the Top-$k$ boundary). The threshold 
$\tau$ is the lower edge of the highest bucket whose suffix mass covers 
the residual target $P_{\text{tgt}} - P_{\text{above}}$, where 
$P_{\text{above}}$ is the mass of blocks scoring above the Top-$k$ 
boundary, found via reverse cumulative summation. A final phase (in the 
full algorithm, Appendix~\ref{label_appendix_B}) atomically allocates 
the selected blocks from a free pool and returns their physical IDs 
($-1$ for skipped blocks, compatible with the paged attention block 
table format). Throughout the paper, especially in evaluation section~\ref{eval_section}, rows marked \textit{(token)} use a token-level variant of both kernels (Algorithm~\ref{alg:token_topp}, Appendix~\ref{label_appendix_B}), with per-token scores instead of block LSEs and no Top-$k$ fast path or forced sink and local-window tokens.

Table~\ref{table:int4_vs_fp16_top_p_comp} compares this approximation 
against an FP16 ground truth (exact sort-and-cumsum over unquantized 
softmax probabilities) on LongBenchV2~\cite{bai2024longbench2}. The 
kernel achieves $\geq 97\%$ recall across all thresholds while matching 
or slightly exceeding FP16 mass coverage, confirming INT4 + histogram 
selection as a high-fidelity drop-in replacement for exact FP16 top-$p$.

\begin{wraptable}[13]{r}{0.50\textwidth}
\centering
\caption{Selection difference between \papertitle{} applied on INT4 vs.\ exact top-\textit{p} selection on FP16 with PyTorch~\cite{bsantraigi_top_p} on LongBenchV2 samples. Mass: average cumulative probability of selected blocks.}
\label{table:int4_vs_fp16_top_p_comp}
\resizebox{0.90\linewidth}{!}{%
\tiny
\begin{tabular}{lcccccc}
\toprule
& \multicolumn{3}{c}{\textbf{Blocks}} & & \multicolumn{2}{c}{\textbf{Mass}} \\
\cmidrule(lr){2-4} \cmidrule(lr){6-7}
Threshold & FP16 & INT4 & Extra & Recall & FP16 & INT4 \\
\midrule
$p=0.8$  & 66.9  & 72.3  & +5.3 & 0.9754 & 0.8346 & 0.8612 \\
$p=0.85$ & 89.1  & 94.9  & +5.8 & 0.9743 & 0.8737 & 0.8906 \\
$p=0.9$  & 123.3 & 129.2 & +5.8 & 0.9723 & 0.9140 & 0.9227 \\
$p=0.95$ & 187.4 & 193.5 & +6.0 & 0.9731 & 0.9559 & 0.9590 \\
\bottomrule
\end{tabular}}
\vspace{-10pt}
\end{wraptable}

\paragraph{Token Relocation Guided by Accumulated Prefill Attention.}
\label{para:relocation}

Algorithm~\ref{alg:token_relocation} details the token relocation 
procedure for mitigating \textit{intra-block sparsity}. After prefill, 
tokens reside in chronological order in the KV cache. For block-sparse 
attention during decode, this ordering is suboptimal: tokens with high 
attention scores, those critical for reaching threshold $p$, may be 
scattered across many blocks, forcing top-$p$ to load far more blocks 
than necessary (Figure~\ref{fig:intra_block_sparsity}). We address this 
by physically rearranging KV cache entries after prefill, placing the 
most attended tokens into the earliest blocks so that top-$p$ reaches 
the same $p$ with fewer blocks. This is safe because RoPE~\cite{su2024roformer} is applied to keys before caching, so positional information is already encoded in the keys after being cached. Importance is measured by the total 
attention each token receives during prefill 
(line~\ref{alg:line:prefill_accum}), summed across the 
$G = H_q / H_{\text{kv}}$ query heads within each GQA group 
(line~\ref{alg:line:gqa_sum}). Relocation is performed independently 
per KV head, since heads exhibit different attention patterns 
(Figure~\ref{fig:dsa_sparsity_motivation_head}).

Naive sorting by accumulated attention destroys the causal locality 
that attention relies on, eliminating the benefit. Instead, we use 
\emph{banded sorting} (Algorithm~\ref{alg:token_relocation}): tokens 
are ranked globally by attention scores, then partitioned into $R$ 
contiguous importance bands (band~1 most important, band~$R$ least). 
Within each band, tokens are re-sorted by their original position to 
restore chronological order. The K and V caches are then permuted 
accordingly via gather operations per head. The result: early blocks 
concentrate the most attended tokens while preserving sequential 
structure within each band, reducing the number of blocks selected by 
top-$p$ by up to 14\% (\S\ref{eval_section}). Relocation runs once 
per request after prefill completes, with negligible cost relative 
to prefill itself.

%% file: algorithms/token_relocation_gemini_version.tex
\begin{wrapfigure}[20]{r}{0.6\textwidth}
\vspace{-25pt}
\begin{minipage}{\linewidth}
\hrule height 0.8pt
\vspace{4pt}
\refstepcounter{algorithm}
\noindent\textbf{Algorithm \thealgorithm} Importance-Guided Token Relocation
\label{alg:token_relocation}
\vspace{4pt}
\hrule height 0.4pt
\vspace{4pt}
\scriptsize
\linespread{0.9}\selectfont
\begin{algorithmic}[1]
\Require KV caches $\mathbf{K}, \mathbf{V} \in \mathbb{R}^{H_{kv} \times L \times d}$; prefill query/key $\mathbf{Q} \in \mathbb{R}^{H_q \times L \times d}, \hat{\mathbf{K}} \in \mathbb{R}^{H_{kv} \times L \times d}$; bands $R$
\Ensure Caches with important tokens concentrated in early blocks
\Statex \textbf{Phase 1: Accumulate prefill attention scores}
\State $\mathbf{A} \gets \mathbf{Q} \, \hat{\mathbf{K}}^\top / \sqrt{d}$ \Comment{Scaled dot-product with causal masking}
\State $\mathbf{S}_h \gets \sum_{t=1}^{L} \mathbf{A}_{h,t,:} \quad \forall\, h \in [H_q]$ \Comment{Sum over query positions} \label{alg:line:prefill_accum}
\Statex \textbf{Phase 2: Aggregate importance per KV head}
\For{$h = 1$ \textbf{to} $H_{kv}$}
    \State $\mathbf{s}_h \gets \sum_{g=1}^{G} \mathbf{S}_{(h-1)G+g}$ \Comment{Pool $G = H_q / H_{kv}$ grouped heads} \label{alg:line:gqa_sum}
\EndFor
\Statex \textbf{Phase 3: Rank and partition into importance bands}
\For{$h = 1$ \textbf{to} $H_{kv}$}
    \State $\boldsymbol{\pi}_h \gets \mathrm{argsort}(\mathbf{s}_h,\;\texttt{descending})$ \Comment{Rank $L$ tokens}
    \State Partition top $R \cdot d_{\text{band}}$ into $R$ contiguous bands of size $d_{\text{band}}$;
           leave the tail in a single trailing band
    \For{each importance band $r = 1, \dots, R$}
        \State Sort indices within band $r$ in \emph{ascending} order
    \EndFor
    \State Sort indices in the tail band in \emph{ascending} order
\EndFor
\Statex \textbf{Phase 4: Reorder caches}
\For{$h = 1$ \textbf{to} $H_{kv}$}
    \State $\mathbf{K}_h \gets \mathrm{gather}(\mathbf{K}_h,\;\boldsymbol{\pi}_h)$; $\mathbf{V}_h \gets \mathrm{gather}(\mathbf{V}_h,\;\boldsymbol{\pi}_h)$
\EndFor
\end{algorithmic}
\vspace{4pt}
\hrule height 0.8pt
\end{minipage}
\vspace{-1.5em}
\end{wrapfigure}

%% file: evaluation.tex
\section{Evaluation}
\label{eval_section}

\begin{wraptable}[11]{r}{0.55\textwidth}
\centering
\vspace{-30pt}
\caption{Average scores on 12 tasks from LongBench. Budget is the average number of selected tokens per layer/head. Detailed per-dataset results are in Table~\ref{tab:longbench} (Appendix~\ref{label_appendix_A}). DS results are from~\cite{lin2025twilight,yang2024double}.}
\label{tab:longbench_summary}
\renewcommand{\arraystretch}{1.1}
\resizebox{0.50\textwidth}{!}{%
\begin{tabular}{@{}lcccc@{}}
\toprule
& \multicolumn{2}{c}{\textbf{LongChat-7B}} & \multicolumn{2}{c}{\textbf{LLaMA-3.1-8B}} \\
\cmidrule(lr){2-3} \cmidrule(lr){4-5}
\textbf{Method} & \textbf{Budget} & \textbf{Avg.} & \textbf{Budget} & \textbf{Avg.} \\
\midrule
Full Attention & 32k & 36.10 & 32k & 52.01 \\
\midrule
Quest+Twi & 137 & 37.33 {\scriptsize\textcolor{ForestGreen}{+3.4\%}} & 489 & 52.00 {\scriptsize\textcolor{red}{-0.0\%}} \\
DS+Twi    & 126 & 37.98 {\scriptsize\textcolor{ForestGreen}{+5.2\%}} & 446 & 51.73 {\scriptsize\textcolor{red}{-0.5\%}} \\
\midrule
\rowcolor{green!6}
Quest+\papertitle{} & 141 & 37.66 {\scriptsize\textcolor{ForestGreen}{+4.3\%}} & 520 & 52.10 {\scriptsize\textcolor{ForestGreen}{+0.2\%}} \\
\rowcolor{green!6}
\papertitle{}       & 170 & \textbf{37.79} {\scriptsize\textcolor{ForestGreen}{+4.7\%}} & 610 & \textbf{52.20} {\scriptsize\textcolor{ForestGreen}{+0.4\%}} \\
\bottomrule
\end{tabular}%
}
\vspace{-10pt}
\end{wraptable}

We evaluate \papertitle{} on 6 different models: Llama 3.1-8B-Instruct\cite{dubey2024llama}, LLaMA2-7B-Chat\cite{touvron2023llama}, Longchat-7B-v1.5-32k\cite{li2023longchat}, Qwen2.5-7B-Instruct~\cite{yang2025qwen25}, Qwen3-32B~\cite{yang2025qwen3}, and Phi-3-mini-128k-instruct\cite{abdin2024phi3} based on 2 parts: (1) Accuracy and kernel performance (\S\ref{acc_perf_section}) on NVIDIA A100, RTX 6000 Ada, and RTX PRO 6000 Blackwell GPUs; (2) System performance (\S\ref{system_section}), on different benchmarks using 2 A100-40GBs, one for prefill and another for decode engines. Benchmarks mainly include long-context understanding (LongBench~\cite{bai2024longbench2,bai2024longbench}), retrieval (RULER~\cite{hsieh2024ruler}), medium-context datasets: GSM8K~\cite{cobbe2021training}, COQA\cite{reddy2019coqa} and PG-19's perplexity test\cite{rae2019compressive}.

\subsection{Accuracy and Kernel Performance.}
\label{acc_perf_section}
\textbf{Benchmarks}. For this experiment, we evaluate \papertitle{} for its accuracy and kernel performance. The baselines include different SOTA sparse-attention algorithms, such as Quest\cite{tang2024quest}, DS\cite{yang2024double}, and Twilight\cite{lin2025twilight}, and full/dense attention. Following the baselines, we do not apply any sparse methods to the first two layers to ensure a fair comparison. For DS, we use the optimized configurations tuned for each model provided by its official repository. The threshold $p$ of Twilight and \papertitle{} is set to 0.95 for LLaMA-2/3, and 0.85 for Longchat ~\cite{lin2025twilight} to align with other works. For all baselines, we follow their official repository.

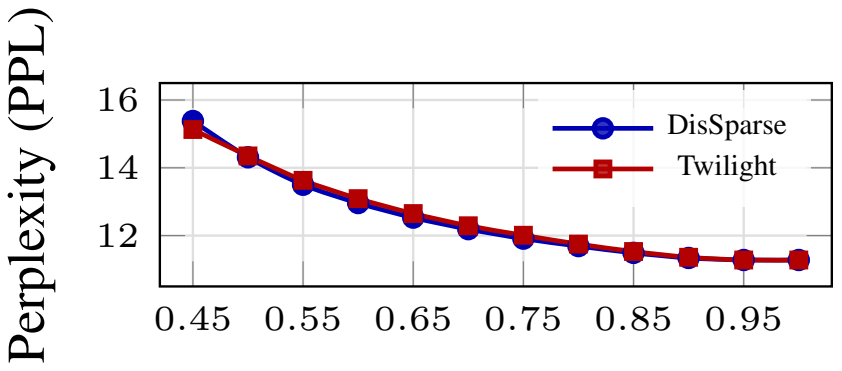
\begin{wrapfigure}{r}{0.40\textwidth}
\centering
\vspace{-20pt}
\resizebox{0.40\textwidth}{!}{%
\begin{tikzpicture}
\begin{axis}[
    width=0.40\textwidth,
    height=0.20\textwidth,
    ylabel={Perplexity (PPL)},
    xmin=0.42, xmax=1.03,
    ymin=10.5, ymax=16.5,
    xtick={0.45,0.55,0.65,0.75,0.85,0.95},
    xticklabel style={/pgf/number format/fixed, /pgf/number format/precision=2, font=\tiny},
    yticklabel style={font=\tiny},
    grid=major,
    grid style={gray!25},
    tick label style={font=\tiny},
    label style={font=\small},
    legend style={
        at={(0.97,0.97)}, anchor=north east,
        font=\fontsize{5}{6}\selectfont,
        draw=none, fill=white, fill opacity=0.8, text opacity=1,
        row sep=-2pt, inner sep=2pt,
    },
]
\addplot[
    color=blue!70!black,
    mark=*,
    mark size=1.5pt,
    thick,
    smooth,
] coordinates {
    (0.45, 15.38) (0.50, 14.31) (0.55, 13.50) (0.60, 12.96)
    (0.65, 12.53) (0.70, 12.19) (0.75, 11.91) (0.80, 11.69)
    (0.85, 11.49) (0.90, 11.34) (0.95, 11.28) (1.00, 11.28)
};
\addlegendentry{\papertitle{}}
\addplot[
    color=red!70!black,
    mark=square*,
    mark size=1.2pt,
    thick,
    smooth,
] coordinates {
    (0.45, 15.13) (0.50, 14.35) (0.55, 13.63) (0.60, 13.09)
    (0.65, 12.65) (0.70, 12.29) (0.75, 12.01) (0.80, 11.75)
    (0.85, 11.53) (0.90, 11.36) (0.95, 11.28) (1.00, 11.28)
};
\addlegendentry{Twilight}
\end{axis}
\end{tikzpicture}}
\caption{Perplexity on PG-19 vs.\ top-$p$ threshold for LLaMA-3.1-8B.}
\label{fig:perplexity_sweep_pg19}
\end{wrapfigure}

\paragraph{Accuracy Results on LongBench. }Table \ref{tab:longbench} summarizes the accuracy result of Twilight\cite{lin2025twilight} integrated with Quest\cite{tang2024quest} and DS\cite{yang2024double}, dense attention, and \papertitle{} on 12 datasets from LongBench dataset\cite{bai2024longbench}. 
Across both models, \papertitle{} matches or exceeds dense attention accuracy. On LongChat-7B-v1.5-32k, \papertitle{} improves by up to +4.7\% over the dense baseline at a budget of only $~170$ tokens, likely because adaptive top-$p$ selection filters low-attention noise that contributes to output degradation. On LLaMA-3.1-8B-Instruct, \papertitle{}, and the baselines all land within ±$0.5\%$ of the dense baseline, within the natural variance of LongBench metrics. The key result is that \papertitle{} achieves this with substantially smaller selection budgets than dense attention requires, similar to Twilight baselines. \papertitle{} also outperforms Ada-KV~\cite{feng2025adakv} at similar budgets (Table~\ref{tab:adakv}).

\begin{wraptable}[15]{r}{0.46\textwidth}
\centering
\vspace{-15pt}
\footnotesize
\caption{Results for GSM8K, COQA, PG-19 (length 4K for LLaMA-2, 16K for LLaMA-3.1). Token Budget: Twilight/\papertitle{}.}
\label{tab:medium_context}
\resizebox{0.45\textwidth}{!}{%
\begin{tabular}{@{}lccc@{}}
\toprule
 & GSM8K(flexible/strict)$\uparrow$ & COQA(em/f1)$\uparrow$ & PG-19 PPL$\downarrow$ \\
\midrule
\multicolumn{4}{c}{LLaMA-2-7B-Chat} \\
\midrule
Full (dense)  & 0.227/0.225 & 0.606/0.761 & \textbf{10.85} \\
Quest-128     & 0.052/0.051 & 0.571/0.744 & 27.47 \\
Twilight      & \textbf{0.225/0.225} & \textbf{0.599/0.759} & 11.09 \\
\papertitle{} & 0.223/0.221 & 0.597/0.758 & 11.07 \\
Token Budget        & \textit{105/117} & \textit{112/125} & \textit{415/452} \\
\midrule
\multicolumn{4}{c}{LLaMA-3.1-8B-Instruct} \\
\midrule
Full (dense)  & 0.776/0.713 & 0.642/0.784 & \textbf{11.28} \\
Quest-128     & 0.401/0.384 & 0.611/0.760 & 79.23 \\
Twilight      & \textbf{0.785/0.737} & 0.639/0.781 & 11.28 \\
\papertitle{} & 0.779/0.726 & \textbf{0.643/0.784} & 11.28 \\
Token Budget        & \textit{89/97} & \textit{93/103} & \textit{1187/1230} \\
\bottomrule
\end{tabular}}
\vspace{-10pt}
\end{wraptable}

\paragraph{Accuracy Results on medium-length context.}
Table~\ref{tab:medium_context} reports accuracy on
GSM8K~\cite{cobbe2021training}, COQA~\cite{reddy2019coqa}, and PG-19
perplexity~\cite{rae2019compressive}. \papertitle{} matches Twilight
and dense attention to within about 2\% on both models
across all metrics, while Quest-128 (fixed top-$k$) collapses on
GSM8K (0.40 vs.\ dense 0.78) and PG-19 (PPL 79.23 vs.\ 11.28) on
LLaMA-3.1, confirming that fixed-budget top-$k$ fails on diffuse
attention while top-$p$ adapts. \papertitle{}'s selection budget is
within 12\% of Twilight's (Table~\ref{tab:medium_context}, italic rows).
Figure~\ref{fig:perplexity_sweep_pg19} sweeps top-$p$ from 0.45 to 1.0
on LLaMA-3.1: \papertitle{} tracks Twilight across the range
(gap $\leq 0.25$~PPL), and both converge to within 0.05~PPL
of dense at $p \geq 0.95$. To match Twilight, \papertitle{} runs at
token-level granularity for this experiment.

\paragraph{Accuracy Results on Retrieval.}
A sparse method can fail on retrieval by dropping a rare but critical token. We therefore evaluate 11 RULER~\cite{hsieh2024ruler} subtasks on Llama-3.1-8B-Instruct at 16K and 32K. All sparse configurations stay within 0.8 points of full attention on average, single-needle retrieval is 100\% for \papertitle{} and full attention at both lengths, and \papertitle{} matches Twilight to within 0.4 points (Table~\ref{tab:ruler}, Appendix~\ref{label_appendix_A}).

\begin{figure}[t]
    \centering
    \includegraphics[width=0.99\columnwidth]{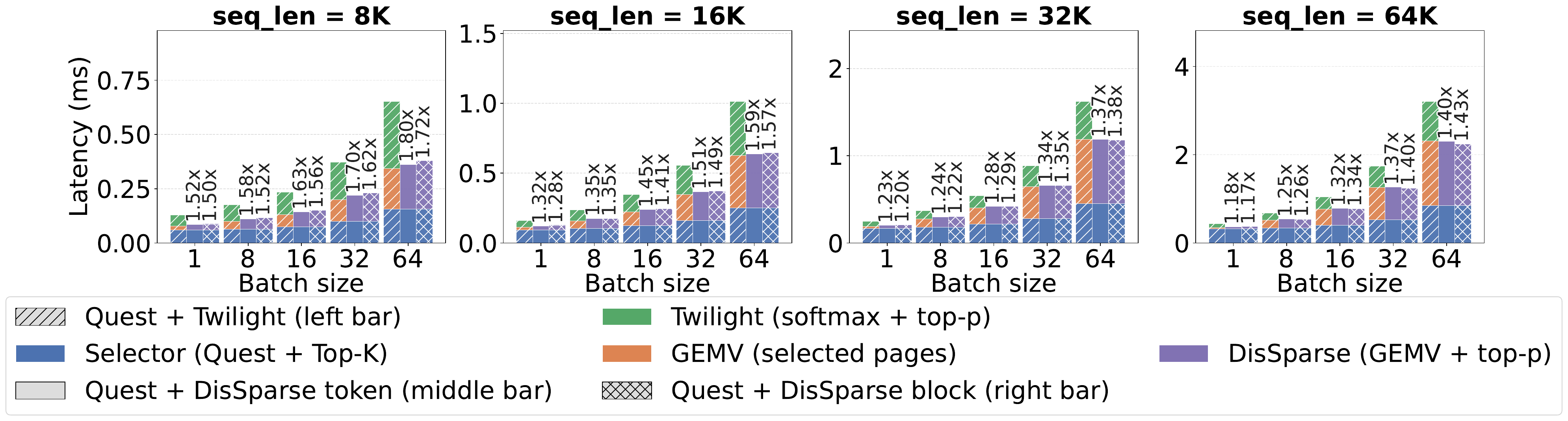}
    \caption{Sparsity analyzer latency for Quest+Twilight vs Quest+DisSparse. DisSparse replaces Twilight's softmax + binary-search post-pass with more efficient GEMV for score calculation and top-$p$ kernel in Algorithm ~\ref{alg:fused_attn_short}.}
    \label{fig:kernel_end_to_end_comp_sparsity}
\end{figure}

\paragraph{Sparsity Analyzer/Selection Performance.} Figure~\ref{fig:kernel_end_to_end_comp_sparsity} isolates the algorithmic contribution: both methods are run in matching two-kernel decompositions (GEMV + top-$p$ post-pass), with Quest-style block selection as the first stage. We compare the end-to-end sparse analyzer latency of Twilight and \papertitle{} and focus on the sparsity analyzer because, in our two-stream pipeline, this component dominates the selection-stream latency and bounds the achievable overlap with attention computation as batch size and sequence length grow (Figure~\ref{fig:layer_breakdown}). 
\papertitle{} speeds up selection by 1.17$\times$ to 1.80$\times$ across all configurations. The speedup grows with batch size (more post-pass parallelism) and is largest at short sequences: at 8K it reaches $1.80\times$ (token level) and $1.72\times$ (block level) at batch~64, while at 64K it ranges from $1.17\times$ to $1.43\times$ as GEMV takes a larger share of the latency and leaves less post-pass time for DisSparse to optimize. \papertitle{}'s algorithms are implemented in CUDA and OpenAI Triton~\cite{tillet2019triton} based on the design details in Section~\ref{solution}.

\paragraph{Token Relocation.}
Table~\ref{tab:reloc_comparison} summarizes the impact of token 
relocation across three models and two long-context benchmarks (more 
details in Tables~\ref{tab:detailed_relocation_results_llama}, 
\ref{tab:detailed_relocation_results_qwen}, and 
\ref{tab:detailed_relocation_results_longchat} in 
Appendix~\ref{label_appendix_A}). Relocation reduces the average 
critical block count by 7.2--14.1\% across the eight (model, 
benchmark, block size) combinations we tested, with the largest 
reductions on LongChat-7B (14.1\%, MHA) and Llama-3.1-8B at block 
size 32 (13.2\%, GQA ratio 4). Accuracy is preserved: on LongBench 
the largest drop is $-0.80$ points (within run-to-run variance), 
and on LongBenchV2 relocation maintains or improves accuracy 
($0$ to $+2.25$ points across Llama-3.1-8B and Qwen2.5-7B). We use 
a band size of $d_{\text{band}} = 4096$ tokens (yielding 
$R = \lceil sequence\_length / d_{\text{band}} \rceil$ bands per sequence) in 
Algorithm~\ref{alg:token_relocation}, which performs well across 
our tested workloads.

\textbf{}

\begin{wraptable}{r}{0.48\textwidth}
\centering
\vspace{-25pt}
\tiny
\caption{Reduction: percentage decrease in average critical block counts ($p$=0.95) with \papertitle{}. Acc. $\Delta$ shows the change in average benchmark score from No Reloc.\ to Reloc.\ (negative: relocation lost accuracy).}
\label{tab:reloc_comparison}
\begin{tabular}{@{}llccc@{}}
\toprule
\textbf{Model} & \textbf{Benchmark} & \textbf{\makecell{Block\\Size}} & \textbf{\makecell{Mean\\Reduction}} & \textbf{Acc.\ $\Delta$} \\
\midrule
\multirow{4}{*}{Llama-3.1-8B}
  & LongBench    & 16 & 11.4\% & $-0.17$ \\
  & LongBench    & 32 & 13.2\% & $-0.44$ \\
  & LongBenchV2  & 16 & 9.4\%  & $+0.79$      \\
  & LongBenchV2  & 32 & 12.4\%     & $0.00$      \\
\midrule
\multirow{3}{*}{Qwen2.5-7B}
  & LongBench    & 16 & 7.9\%  & $-0.17$ \\
  & LongBench    & 32 & 7.6\%  & $-0.80$ \\
  & LongBenchV2  & 16 & 7.2\%  & $+2.25$      \\
\midrule
\multirow{1}{*}{LongChat-7B}
  & LongBench    & 16 & 14.1\% & $-0.22$ \\
\bottomrule
\end{tabular}
\vspace{-10pt}
\end{wraptable}

\vspace{-20pt}

\subsection{System Performance}
\label{system_section}
\paragraph{Setup.}
We integrate \papertitle{} into vLLM with disaggregated inference and evaluate on LongBenchV2 samples up to 32K tokens, using two A100-40GB GPUs for prefill and decode. The decode engine's memory is partitioned among model weights, KV cache (FP16 primary + INT4 replica), and an NCCL buffer that holds the KV cache during asynchronous transfer from the prefill engine. For dense vLLM, the buffer must accommodate FP16 KV caches for the entire batch, whereas \papertitle{} transfers only the INT4 key replica during the first decode step, 8$\times$ smaller, and offloads the FP16 cache to CPU memory in the background. Subsequent decode steps fetch only the blocks selected by the sparse analyzer.

\begin{figure}[htb]
    \centering
    \begin{subfigure}{0.90\columnwidth}
        \centering
        \includegraphics[width=\linewidth]{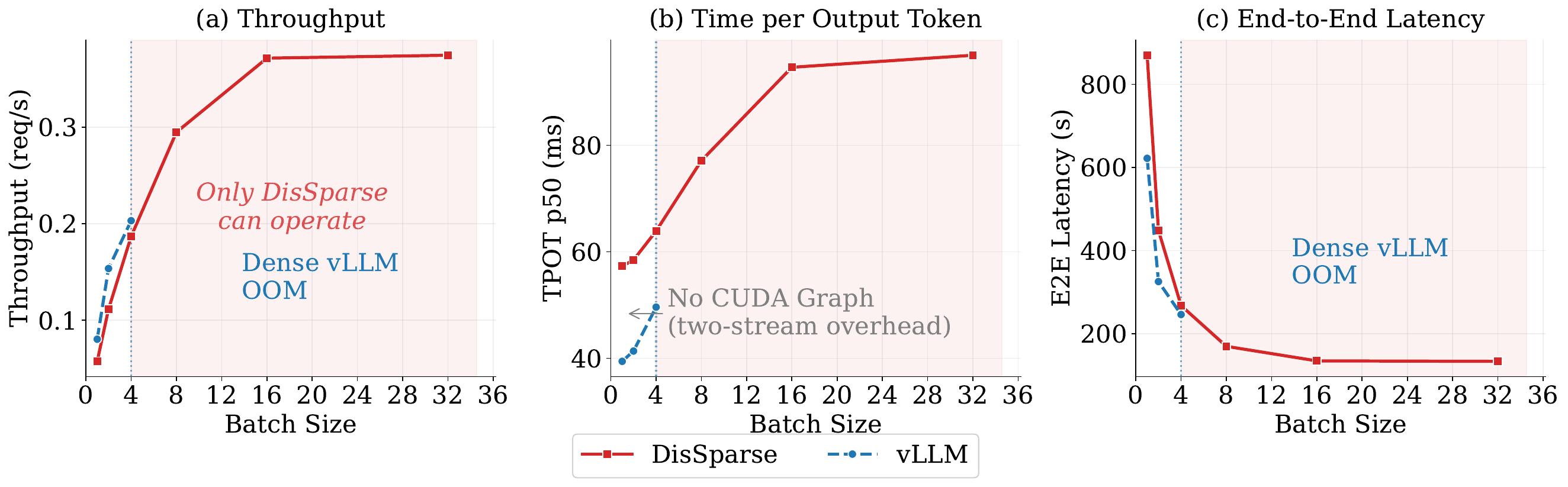}
        \caption{Llama-3.1-8B Performance}
        \label{fig:llama_results}
    \end{subfigure}
    \hfill
    \begin{subfigure}{0.90\columnwidth}
        \centering
        \includegraphics[width=\linewidth]{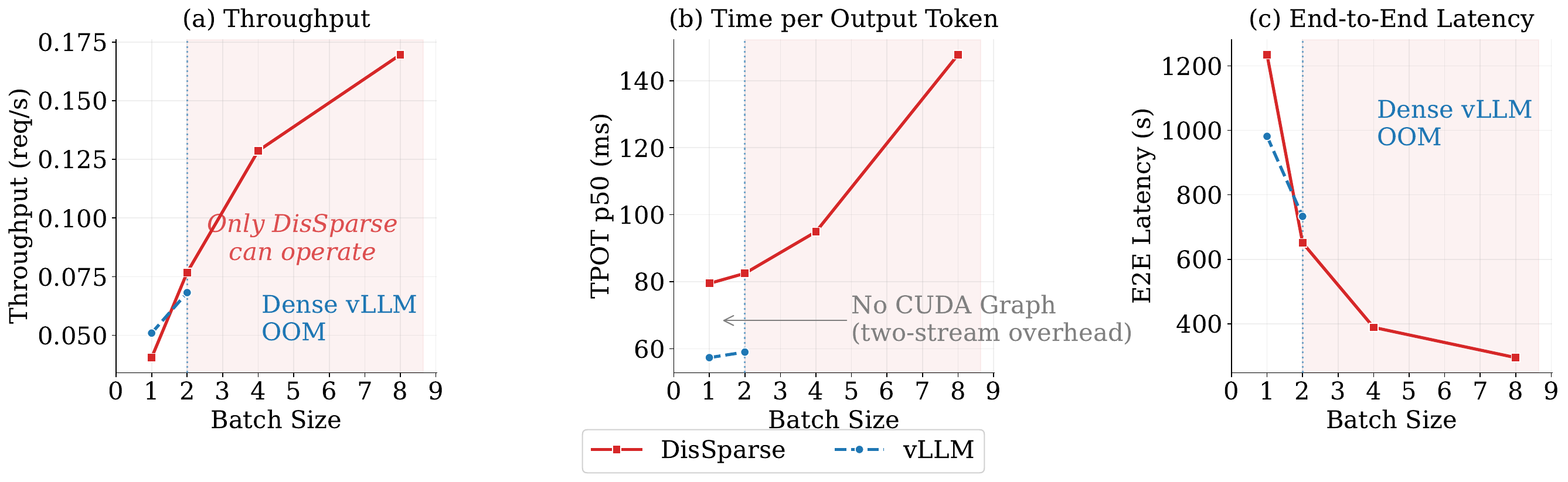}
        \caption{Phi-3-Mini Performance}
        \label{fig:phi_results}
    \end{subfigure}
    \caption{Performance of~\papertitle{} vs. vLLM\cite{kwon2023efficient} with Llama 3.1-8B-Instruct\cite{dubey2024llama} and Phi-3-mini-128k-instruct\cite{abdin2024phi3} models on LongBenchV2 samples\cite{bai2024longbench2} (up to 32k sequence length). Each system is evaluated on 3 metrics: throughput, TPOT, and end-to-end latency vs. batch size.}
    \label{fig:combined_results}
\end{figure}

\paragraph{Results.}
Figure~\ref{fig:combined_results} compares \papertitle{} against dense vLLM on Llama-3.1-8B-Instruct (GQA) and Phi-3-mini-128k-instruct (MHA). Dense vLLM exhausts GPU memory at batch sizes 4 and 2, respectively, while \papertitle{} sustains batch sizes up to 32 and 8 under the same memory budget, achieving 2$\times$ and 2.5$\times$ higher peak throughput. At small batch sizes, dense vLLM is approximately 1.5$\times$ faster in TPOT due to CUDA Graph capture. \papertitle{}'s two-stream pipeline is incompatible with vLLM's single-stream CUDA Graph implementation, so we disable graph capture, exposing the kernel launch overhead. We view this as an engineering limitation rather than a fundamental one\cite{vllm_issue_30211}. Additionally, the implementation in vLLM does not employ Quest as the top-$k$ page filter; hence, \papertitle{} has to do a large GEMV on the whole INT4 cache. Notably, at batch size 2 on Phi-3-mini, \papertitle{} already outperforms dense vLLM on end-to-end latency: the MHA architecture's larger KV cache makes the dense baseline KV-cache-bound at this batch size, while \papertitle{}'s sparse selection reduces the effective KV pressure. The substantial increase in batch size can be related to the reduction in KV cache network traffic from prefill node to decode node. Unlike dense attention, the sparse attention fetches only critical tokens/pages and \papertitle{}'s top-$p$ selection keeps the accuracy high.

%% file: conclusion.tex
\section{Conclusion}
The dominance of top-$k$ in dynamic sparse attention is a consequence of system constraints, not algorithmic merit: fixed budgets align with deterministic memory allocation, but they mismatch the actual variability of attention distributions across heads, layers, and decode steps. We present \papertitle{}, an algorithm-system co-design that integrates top-$p$ selection into vLLM end-to-end by decoupling block selection from the scheduler's critical path, hiding selection latency behind layer computation, and replacing multi-pass selection with a single-pass histogram kernel. Across multiple models and both long- and medium-context benchmarks, \papertitle{} matches dense-attention accuracy while sustaining up to 8$\times$ larger batch sizes than dense vLLM under the same memory budget, evidence that the systems-level barriers to sparse attention can be lifted without algorithmic compromise. The remaining small-batch gap stems from CUDA Graph compatibility, which multi-stream capture should close, and missing Quest as the first layer of sparsity. We discuss the limitations in Appendix~\ref{sec:limitations}.

%% file: acknowledgement.tex
\begin{ack}
We thank the anonymous reviewers for their feedback. This work was partially supported by the National Science Foundation under Grants 2512416, 2339317 (CAREER), and 2448156, and in part by the Commonwealth Cyber Initiative, an investment in the advancement of cyber R\&D, innovation, and workforce development (\url{https://www.cyberinitiative.org}). We also thank Google Cloud Platform and the UCR HPCC for hardware resources.
\end{ack}

%% file: appendix_new_motivation_results.tex
\appendix  
\section{Critical Token Dynamics on Additional Models}
\label{label_appendix_motivation}

\begin{figure}[htbp]
    \centering
    \begin{subfigure}{\textwidth}
        \centering
        \includegraphics[width=\linewidth]{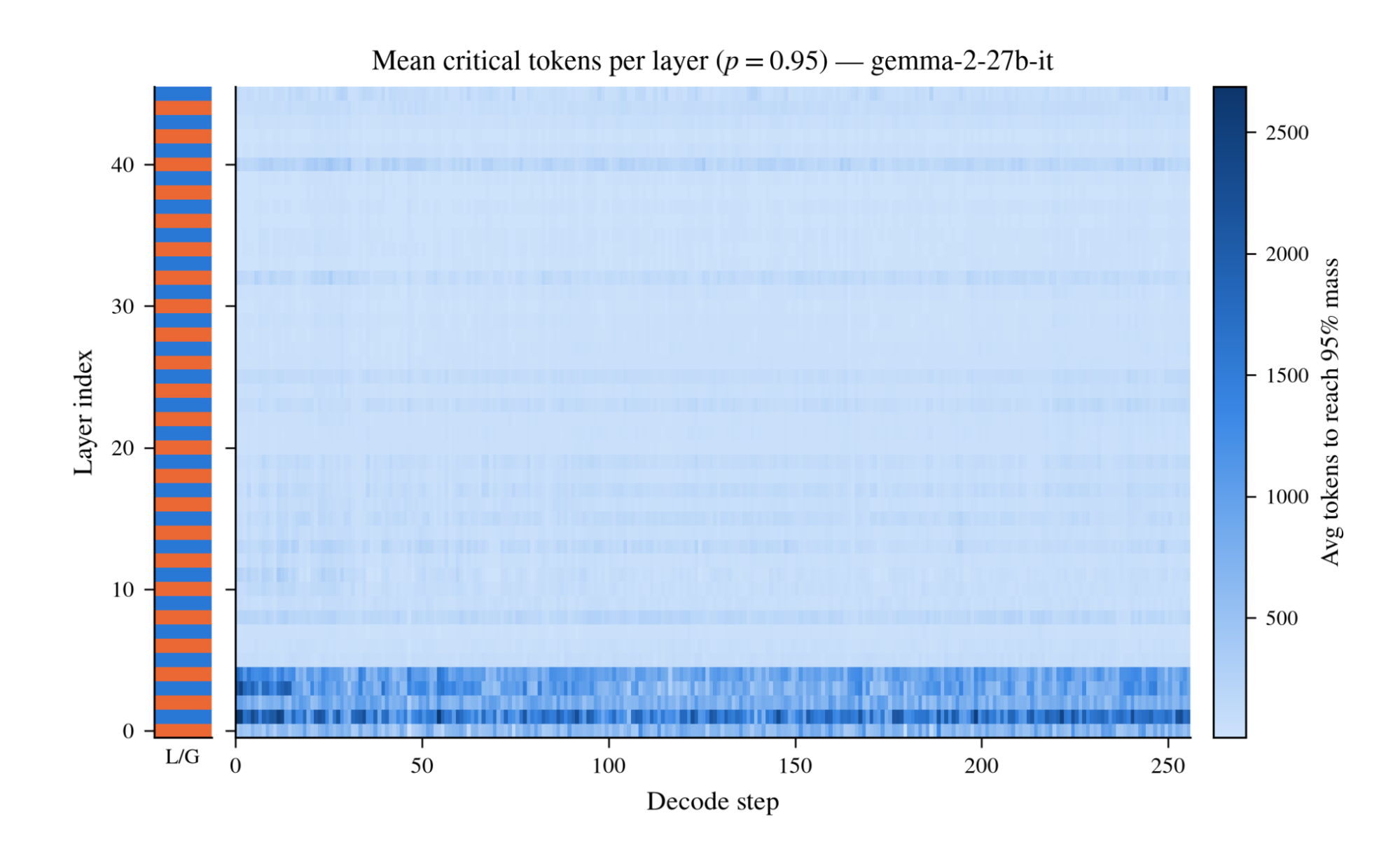}
        \caption{Mean critical tokens per layer. The L/G strip marks local (orange) and global (blue) layers.}
    \end{subfigure}
    \par\medskip
    \begin{subfigure}{\textwidth}
        \centering
        \includegraphics[width=\linewidth]{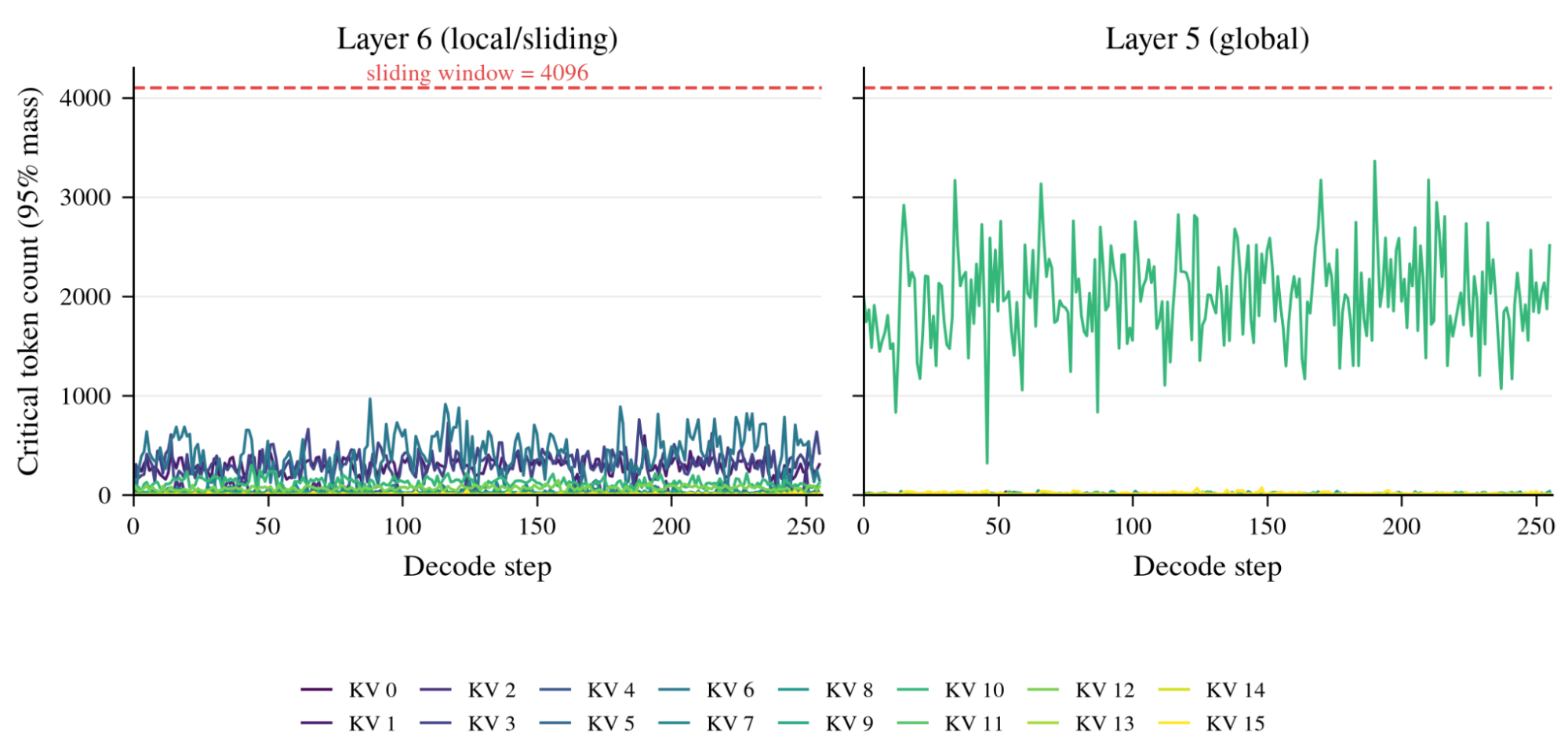}
        \caption{Critical tokens per KV head in a local and a global layer.}
    \end{subfigure}
    \caption{Critical token counts ($p=0.95$) across decode steps for Gemma-2-27B-it.}
    \label{fig:motivation_gemma2}
\end{figure}

In this section of the appendix, Figures~\ref{fig:motivation_qwen25}, \ref{fig:motivation_qwen3}, and \ref{fig:motivation_gemma2} are presented to extend the analysis of Figure~\ref{fig:dsa_sparsity_motivation_head} and \ref{fig:attention_budget} and demonstrate the generality of the observation for \papertitle{}'s motivation, which is the dynamic variation in critical KV cache tokens/blocks. We extend the analysis to Qwen2.5-7B-Instruct~\cite{yang2025qwen25}, Qwen3-32B, and Gemma-2-27B-it with two kinds of plots: per-layer average critical token heatmap and per-KV-head variation in critical token count over 256 decode steps on WikiText samples. Across layers, earliest layers need more tokens compared to later layers (layers 0--3 in Qwen2.5-7B, 0--6 in Qwen3-32B, and 0--4 in Gemma-2-27B) and later layers have fewer but varying average counts of critical tokens. Across KV heads, the critical token count varies significantly. For example, in layer 8 of Qwen3-32B, KV head 7 reaches about 3500 tokens, while KV heads 0, 1, and 6 need a comparatively small set of tokens.

Gemma-2-27B alternates local (sliding-window) and global attention layers, and the two layer types behave differently (Figure~\ref{fig:motivation_gemma2}, bottom). In local layer 6, every KV head stays below about 1000 tokens, well under the 4096-token window. In global layer 5, one KV head needs roughly 1000--3400 tokens at most steps, while the others stay near zero. Together, these results show that the mismatch between a fixed $k$ and attention dynamics described in Section~\ref{problem} extends to these models, including one with a hybrid local and global attention design.

\begin{figure}[htbp]
    \centering
    \begin{subfigure}{\textwidth}
        \centering
        \includegraphics[width=\linewidth]{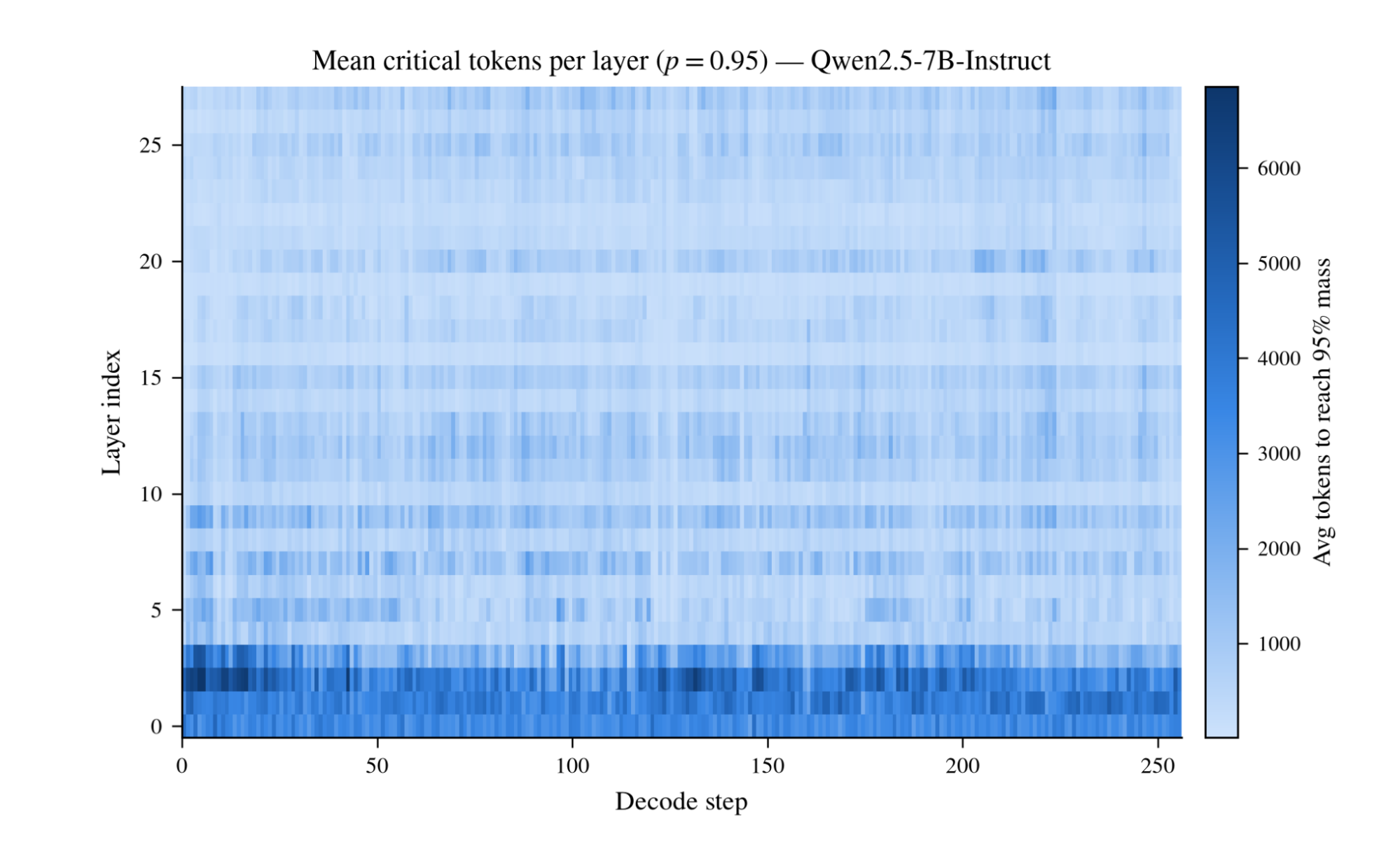}
        \caption{Mean critical tokens per layer.}
    \end{subfigure}
    \par\medskip
    \begin{subfigure}{\textwidth}
        \centering
        \includegraphics[width=\linewidth, trim=0 2 0 0, clip]{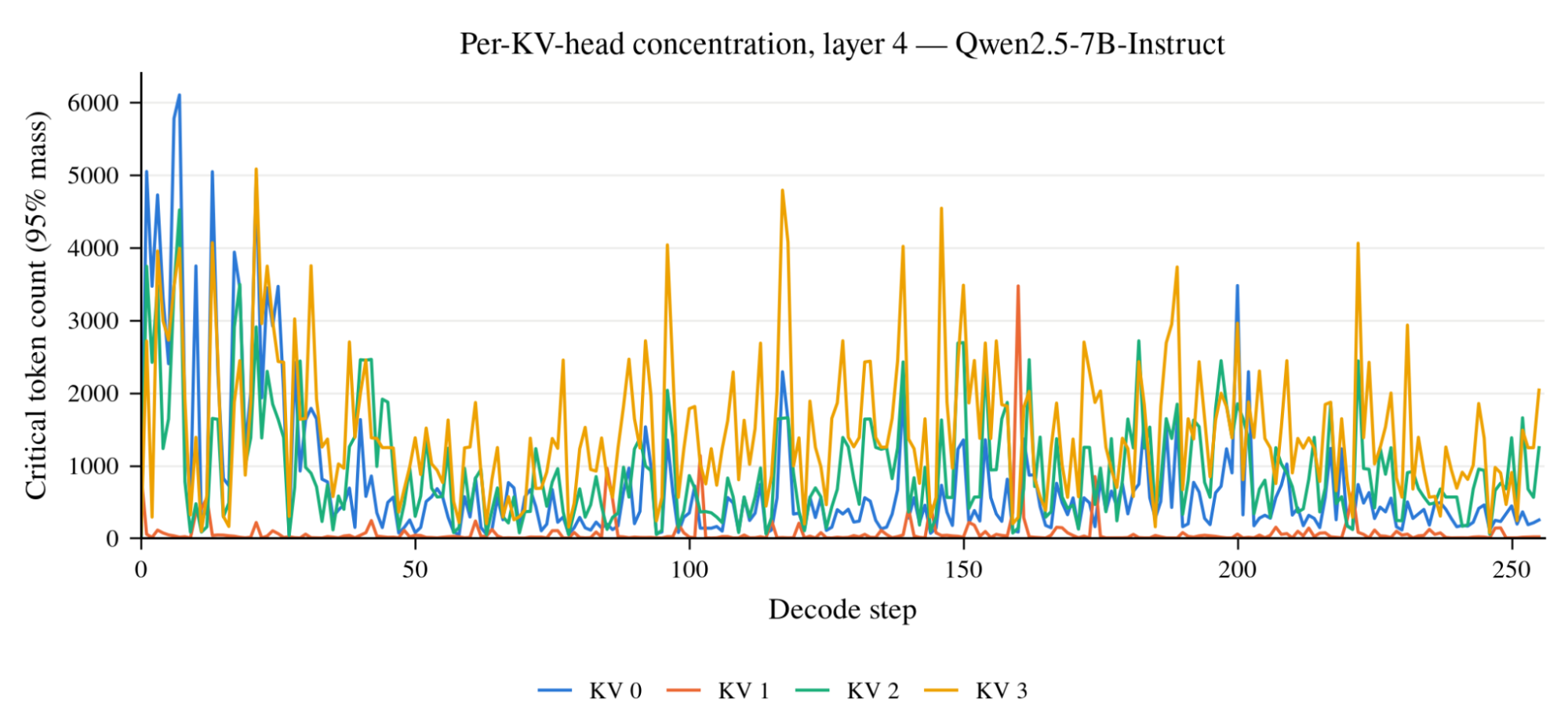}
        \caption{Critical tokens per KV head, layer 4.}
    \end{subfigure}
    \caption{Critical token counts ($p=0.95$) across decode steps for Qwen2.5-7B-Instruct.}
    \label{fig:motivation_qwen25}
\end{figure}

\begin{figure}[htbp]
    \centering
    \begin{subfigure}{\textwidth}
        \centering
        \includegraphics[width=\linewidth]{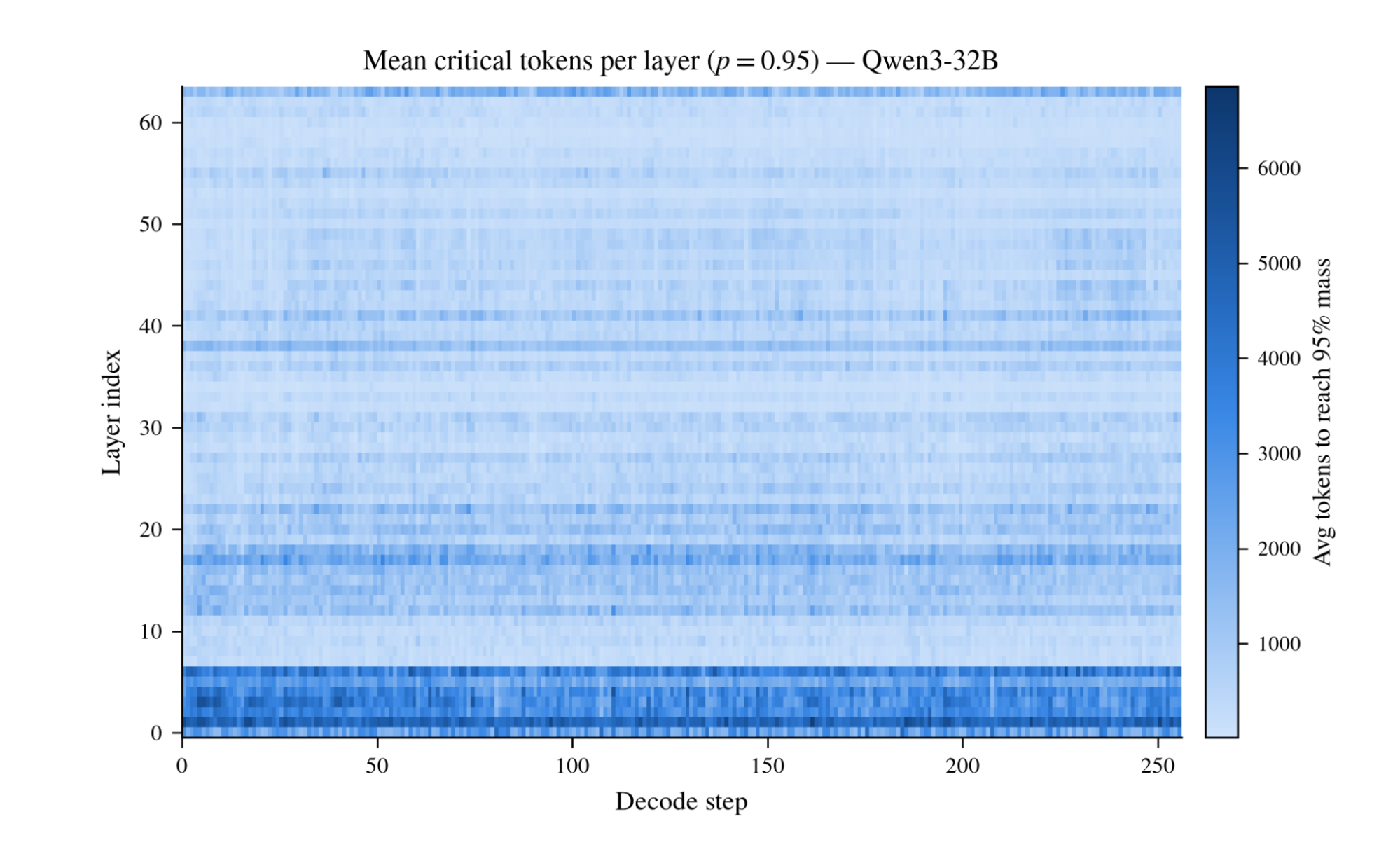}
        \caption{Mean critical tokens per layer.}
    \end{subfigure}
    \par\medskip
    \begin{subfigure}{\textwidth}
        \centering
        \includegraphics[width=\linewidth, trim=0 2 0 0, clip]{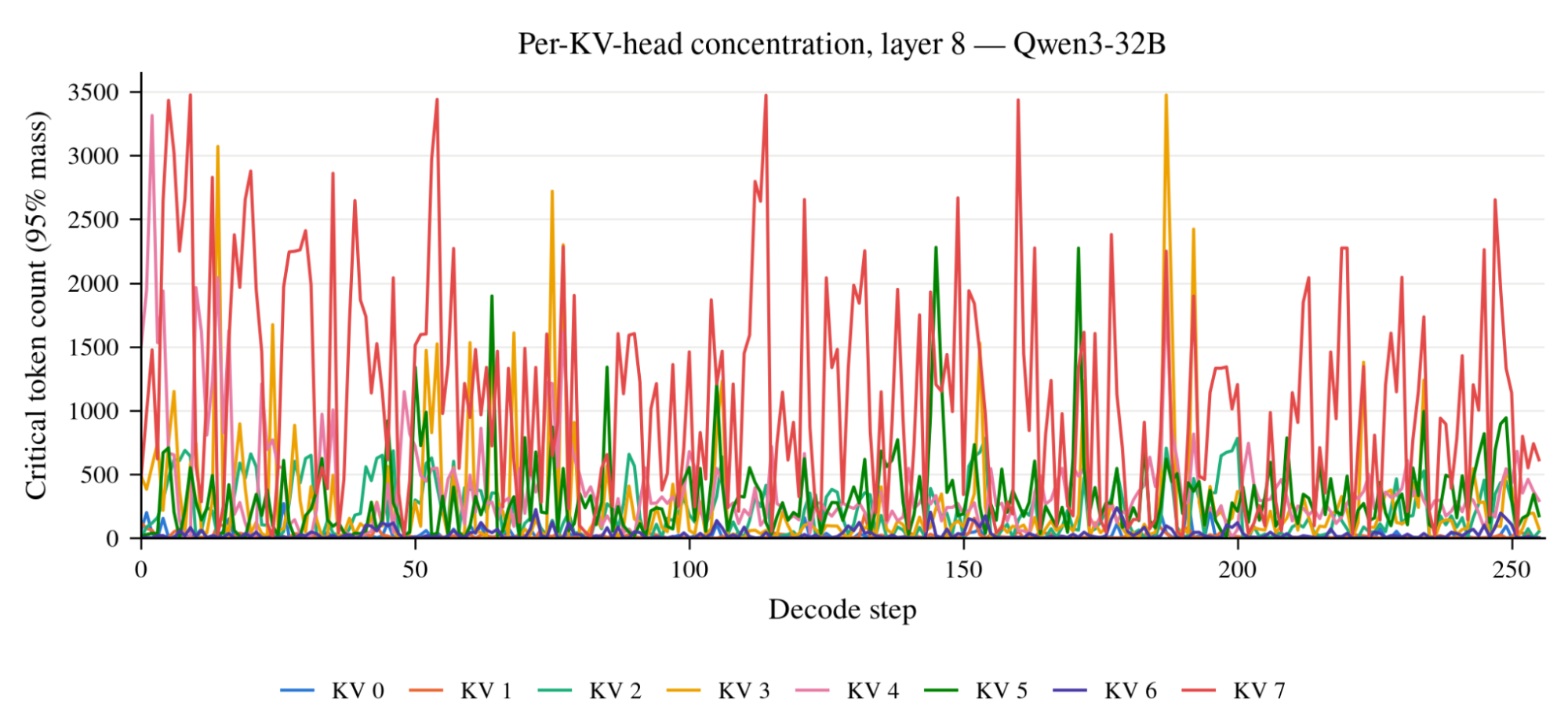}
        \caption{Critical tokens per KV head, layer 8.}
    \end{subfigure}
    \caption{Critical token counts ($p=0.95$) across decode steps for Qwen3-32B.}
    \label{fig:motivation_qwen3}
\end{figure}

\clearpage

%% file: appendix_A.tex
\clearpage
\section{Detailed Results}
\label{label_appendix_A}

\begin{table*}[h]
\centering
\caption{Per-dataset accuracy on LongBench (A100). \textbf{Bold} indicates the best result within each model section (per dataset). \textcolor{ForestGreen}{Green} and \textcolor{red}{red} percentages denote change relative to Full Attention.  Top-$p$ configurations use $p$ = 0.95 for \textit{LlaMA-3.1-8B-Instruct} and $p$ = 0.85 for \textit{LongChat-7B-v1.5-32k}.}
\label{tab:longbench}
\renewcommand{\arraystretch}{1.1}
\resizebox{\textwidth}{!}{%
\begin{tabular}{@{}ll*{2}{c}*{3}{c}*{3}{c}cccc c@{}}
\toprule
& & \multicolumn{2}{c}{\textbf{Single-Doc.\ QA}} & \multicolumn{3}{c}{\textbf{Multi-Doc.\ QA}} & \multicolumn{3}{c}{\textbf{Summarization}} & \textbf{Few} & \textbf{Syn.} & \multicolumn{2}{c}{\textbf{Code}} & \\
\cmidrule(lr){3-4} \cmidrule(lr){5-7} \cmidrule(lr){8-10} \cmidrule(lr){11-11} \cmidrule(lr){12-12} \cmidrule(lr){13-14}
\textbf{Method} & \textbf{Budget} & Qasper & MF-en & HotpotQA & 2WikiMQA & Musique & GovRep. & QMSum & M-News & TriviaQA & PR-en & LCC & Repo. & \textbf{Avg.} \\
\midrule
\multicolumn{15}{c}{\cellcolor{gray!10}\textit{LongChat-7B-v1.5-32k}} \\
\midrule
Full Attn. & 32k & 29.07 & 41.45 & 31.33 & 21.37 & 12.94 & 30.88 & 23.01 & 25.99 & 84.02 & 29.13 & 49.41 & 54.63 & 36.10 \\
\midrule
\multirow{5}{*}{Quest}
  & 256  & 25.64 & 32.31 & 23.50 & 19.93 & 7.36  & 22.53 & 21.20 & 24.95 & 68.03 & 32.10 & 45.66 & 44.27 & 30.62 \\
  & 1024 & 31.19 & 41.69 & 30.83 & 21.98 & 9.98  & 29.79 & 22.94 & 26.29 & 84.99 & 32.96 & 48.31 & 52.99 & 36.16 \\
  & 4096 & 29.36 & 42.04 & 33.33 & 21.55 & 13.07 & 31.38 & 23.10 & 26.35 & 85.15 & 30.09 & 49.86 & 54.53 & 36.65 \\
  & 8192 & 28.94 & 41.04 & 33.66 & 21.12 & 13.34 & 31.03 & 23.26 & 26.38 & 85.49 & 28.66 & 49.71 & 54.58 & 36.43 \\
\rowcolor{blue!6}
  & + Twi.\ {\scriptsize(137)} & 31.51 & 42.60 & 31.99 & 22.39 & 13.31 & 32.05 & 23.03 & 26.23 & 85.72 & 32.00 & 51.44 & 55.69 & 37.33 {\scriptsize\textcolor{ForestGreen}{+3.4\%}} \\
\midrule
\multirow{5}{*}{DS}
  & 256  & 27.89 & 39.15 & 27.42 & 18.68 & 9.22  & 29.53 & 22.02 & 25.59 & 84.75 & 30.57 & 48.77 & 52.49 & 34.67 \\
  & 1024 & 30.13 & 40.62 & 31.21 & 19.69 & 7.18  & 26.69 & 23.19 & 26.41 & 83.99 & 30.09 & 49.91 & 54.51 & 35.30 \\
  & 4096 & 28.55 & 41.24 & 32.90 & 21.29 & 7.97  & 29.53 & 22.99 & 26.45 & 84.12 & 28.66 & 49.48 & 54.49 & 35.64 \\
  & 8192 & 29.05 & 41.42 & 32.16 & 20.66 & 12.34 & 30.29 & 22.74 & 26.33 & 84.41 & 29.13 & 49.57 & 54.34 & 36.04 \\
\rowcolor{blue!6}
  & + Twi.\ {\scriptsize(126)} & 31.89 & 43.20 & \textbf{35.08} & 22.89 & 13.67 & 31.72 & 23.25 & 26.22 & \textbf{86.08} & \textbf{33.91} & \textbf{51.60} & 56.25 & \textbf{37.98} {\scriptsize\textcolor{ForestGreen}{+5.2\%}} \\
\midrule
\rowcolor{blue!6}
Twilight & {\scriptsize Avg.\ 146} & 31.30 & 43.22 & 33.98 & 23.09 & 13.75 & 32.03 & \textbf{23.39} & 26.20 & 85.93 & 32.96 & 51.55 & 56.10 & 37.79 {\scriptsize\textcolor{ForestGreen}{+4.7\%}} \\
\midrule
\rowcolor{green!6}
Quest+\papertitle\ (token) & {\scriptsize Avg.\ 141} & \textbf{32.91} & 42.80 & 33.22 & \textbf{24.80} & 13.78 & 31.73 & 22.68 & \textbf{26.79} & 84.62 & 31.00 & 50.93 & 56.69 & 37.66 {\scriptsize\textcolor{ForestGreen}{+4.3\%}} \\
\rowcolor{green!6}
\papertitle (token) & {\scriptsize Avg.\ 170} & 31.37 & \textbf{43.79} & 34.79 & 23.51 & \textbf{15.00} & \textbf{32.28} & 22.36 & 26.15 & 85.50 & 31.00 & 50.72 & \textbf{56.98} & 37.79 {\scriptsize\textcolor{ForestGreen}{+4.7\%}} \\
\bottomrule
\midrule
\multicolumn{15}{c}{\cellcolor{gray!10}\textit{LLaMA-3.1-8B-Instruct}} \\
\midrule
Full Attn. & 128k & 46.17 & 53.33 & 55.36 & 43.95 & 27.08 & 35.01 & 25.24 & 27.37 & 91.18 & 99.50 & 62.17 & 57.76 & 52.01 \\
\midrule
\multirow{5}{*}{Quest}
  & 256  & 25.20 & 38.10 & 29.10 & 29.10 & 9.30  & 26.50 & 19.70 & 26.50 & 66.20 & 95.50 & 53.00 & 47.10 & 38.80 \\
  & 1024 & 39.90 & 50.90 & 47.50 & 37.40 & 22.90 & 33.70 & 23.80 & \textbf{27.50} & 80.90 & 99.50 & 59.90 & 56.70 & 48.40 \\
  & 4096 & 45.80 & 51.10 & 51.60 & 37.90 & 24.80 & 34.70 & 25.10 & \textbf{27.50} & 89.80 & 99.50 & 63.00 & 57.30 & 50.70 \\
  & 8192 & 45.10 & 53.00 & 54.90 & 44.80 & 25.40 & 34.60 & \textbf{25.60} & 27.30 & 91.40 & 99.50 & 62.90 & 57.50 & 51.80 \\
\rowcolor{blue!6}
  & + Twi.\ {\scriptsize(489)} & 45.94 & 53.15 & 53.93 & 46.04 & 25.10 & 34.20 & 25.35 & 27.20 & 90.65 & 99.50 & 63.00 & 60.20 & 52.00 {\scriptsize\textcolor{red}{-0.0\%}} \\
\midrule
\multirow{5}{*}{DS}
  & 256  & 38.24 & 49.58 & 43.38 & 31.98 & 15.52 & 33.40 & 24.06 & 26.86 & 84.41 & 99.50 & 53.28 & 48.64 & 45.74 \\
  & 1024 & 42.97 & \textbf{54.65} & 51.75 & 33.92 & 20.39 & 34.50 & 24.92 & 26.71 & \textbf{92.81} & 99.50 & 62.66 & 48.37 & 49.43 \\
  & 4096 & 43.50 & 53.17 & 54.21 & 44.70 & 23.14 & \textbf{34.73} & 25.40 & 26.71 & 92.78 & 99.50 & 62.59 & 51.31 & 50.98 \\
  & 8192 & 43.82 & 53.71 & 54.19 & 45.13 & 23.72 & 34.27 & 24.98 & 26.69 & 91.61 & \textbf{100.00} & 62.40 & 52.87 & 51.14 \\
\rowcolor{blue!6}
  & + Twi.\ {\scriptsize(446)} & 43.08 & 52.89 & 54.68 & 44.86 & 24.88 & 34.09 & 25.20 & 27.00 & 91.20 & \textbf{100.00} & \textbf{63.95} & 58.93 & 51.73 {\scriptsize\textcolor{red}{-0.5\%}} \\
\midrule
\rowcolor{blue!6}
Twilight & {\scriptsize Avg.\ 510} & 46.00 & 53.10 & 54.00 & 46.00 & 25.10 & 34.20 & 25.40 & 27.20 & 90.60 & 99.50 & 63.00 & 60.20 & 52.00 {\scriptsize\textcolor{red}{-0.0\%}} \\
\midrule
\rowcolor{green!6}
Quest+\papertitle (token) & {\scriptsize Avg.\ 520} & \textbf{46.60} & 52.50 & 54.80 & \textbf{46.50} & 24.80 & 34.50 & 25.40 & 27.40 & 90.30 & 99.50 & 63.30 & 59.70 & 52.10 {\scriptsize\textcolor{ForestGreen}{+0.2\%}} \\
\rowcolor{green!6}
\papertitle (token) & {\scriptsize Avg.\ 610} & 45.60 & 52.90 & \textbf{55.10} & 44.70 & \textbf{28.00} & 33.90 & 25.20 & 27.20 & 90.60 & 99.50 & 63.00 & \textbf{60.40} & \textbf{52.20} {\scriptsize\textcolor{ForestGreen}{+0.4\%}} \\
\bottomrule
\end{tabular}%
}
\end{table*}

\begin{table*}[h]
\centering
\caption{Per-dataset accuracy on LongBench (RTX 6000 Ada). \textbf{Bold} indicates the best result
within each model section (per dataset). All top-$p$ configurations use $p$ = 0.95.
\textcolor{ForestGreen}{Green} and \textcolor{red}{red}
percentages denote change relative to Full Attention. }
\label{tab:longbench_ada}
\renewcommand{\arraystretch}{1.1}
\resizebox{\textwidth}{!}{%
\begin{tabular}{@{}ll*{2}{c}*{3}{c}*{3}{c}cccc c@{}}
\toprule
& & \multicolumn{2}{c}{\textbf{Single-Doc.\ QA}} & \multicolumn{3}{c}{\textbf{Multi-Doc.\ QA}} & \multicolumn{3}{c}{\textbf{Summarization}} & \textbf{Few} & \textbf{Syn.} & \multicolumn{2}{c}{\textbf{Code}} & \\
\cmidrule(lr){3-4} \cmidrule(lr){5-7} \cmidrule(lr){8-10} \cmidrule(lr){11-11} \cmidrule(lr){12-12} \cmidrule(lr){13-14}
\textbf{Method} & \textbf{Budget} & Qasper & MF-en & HotpotQA & 2WikiMQA & Musique & GovRep. & QMSum & M-News & TriviaQA & PR-en & LCC & Repo. & \textbf{Avg.} \\
\midrule
\multicolumn{15}{c}{\cellcolor{gray!10}\textit{LongChat-7B-v1.5-32k}} \\
\midrule
Full Attn. & 32k & 27.23 & 41.45 & 30.45 & 23.53 & \textbf{13.02} & 31.42 & 22.69 & 26.52 & 83.96 & 31.50 & 54.65 & 58.04 & 37.04 \\
\midrule
\multirow{5}{*}{Quest}
  & 256  & 23.31 & 33.44 & 20.15 & 21.68 & 7.39 & 22.88 & 21.39 & 25.79 & 69.62 & \textbf{34.50} & 50.17 & 48.56 & 31.57 {\scriptsize\textcolor{red}{-14.8\%}} \\
  & 1024  & 28.91 & 40.89 & 30.62 & 22.27 & 11.41 & 31.18 & 22.22 & 26.46 & \textbf{84.72} & 34.50 & 52.21 & 57.42 & 36.90 {\scriptsize\textcolor{red}{-0.4\%}} \\
  & 4096  & 28.68 & 42.18 & 30.31 & 23.08 & 11.19 & 31.56 & 22.26 & \textbf{26.69} & 83.97 & 29.00 & 54.70 & 58.47 & 36.84 {\scriptsize\textcolor{red}{-0.5\%}} \\
  & 8192  & 27.47 & 41.36 & 31.28 & 23.65 & 11.71 & 31.33 & 22.36 & 26.38 & 83.91 & 28.00 & 54.41 & 57.93 & 36.65 {\scriptsize\textcolor{red}{-1.1\%}} \\
\rowcolor{blue!6}
  & + Twi.\ {\scriptsize(398)} & \textbf{29.48} & \textbf{42.85} & 31.16 & \textbf{24.11} & 12.36 & 31.74 & 22.56 & 26.45 & 84.49 & 30.50 & 54.74 & \textbf{58.73} & 37.43 {\scriptsize\textcolor{ForestGreen}{+1.1\%}} \\
\rowcolor{green!6}
  & + \papertitle\ (token) {\scriptsize(424)} & 29.11 & 42.60 & 31.49 & 23.84 & 12.73 & 31.77 & 22.61 & 26.30 & 84.02 & 32.00 & 54.61 & 58.38 & \textbf{37.45} {\scriptsize\textcolor{ForestGreen}{+1.1\%}} \\
\rowcolor{green!6}
  & + \papertitle\ (block) {\scriptsize(982)} & 28.94 & 40.34 & \textbf{32.57} & 23.93 & 12.35 & 31.72 & \textbf{22.76} & 26.54 & 83.97 & 33.00 & 53.89 & 58.52 & 37.38 {\scriptsize\textcolor{ForestGreen}{+0.9\%}} \\
\midrule
\rowcolor{blue!6}
Twilight & {\scriptsize Avg.\ 468} & 28.66 & 41.44 & 31.26 & 23.41 & 12.78 & \textbf{32.01} & 22.36 & 26.45 & 83.54 & 33.00 & 55.25 & 58.60 & 37.40 {\scriptsize\textcolor{ForestGreen}{+1.0\%}} \\
\rowcolor{green!6}
\papertitle (token) & {\scriptsize Avg.\ 500} & 28.46 & 40.40 & 31.31 & 23.47 & 12.58 & 31.71 & 22.59 & 26.47 & 83.81 & 34.50 & \textbf{55.29} & 58.61 & 37.43 {\scriptsize\textcolor{ForestGreen}{+1.1\%}} \\
\midrule\multicolumn{15}{c}{\cellcolor{gray!10}\textit{LLaMA-3.1-8B-Instruct}} \\
\midrule
Full Attn. & 128k & 46.10 & \textbf{52.98} & \textbf{56.17} & 43.65 & \textbf{28.41} & \textbf{34.85} & 25.19 & \textbf{27.42} & 91.47 & \textbf{99.50} & 64.51 & 59.60 & 52.49 \\
\midrule
\multirow{5}{*}{Quest}
  & 256  & 24.71 & 36.54 & 30.73 & 28.81 & 13.51 & 26.56 & 19.58 & 26.45 & 64.96 & 92.00 & 54.43 & 48.38 & 38.89 {\scriptsize\textcolor{red}{-25.9\%}} \\
  & 1024  & 39.59 & 49.91 & 47.65 & 38.96 & 20.66 & 33.42 & 23.85 & 27.29 & 81.53 & 99.50 & 60.66 & 57.52 & 48.38 {\scriptsize\textcolor{red}{-7.8\%}} \\
  & 4096  & \textbf{47.15} & 51.81 & 52.16 & 37.67 & 23.72 & 34.47 & 25.23 & 27.41 & 89.82 & 99.50 & 64.85 & 59.85 & 51.14 {\scriptsize\textcolor{red}{-2.6\%}} \\
  & 8192  & 46.39 & 52.39 & 54.83 & 45.07 & 25.82 & 34.75 & 25.47 & 27.41 & \textbf{91.73} & 99.50 & 64.69 & 59.49 & 52.29 {\scriptsize\textcolor{red}{-0.4\%}} \\
\rowcolor{blue!6}
  & + Twi.\ {\scriptsize(489)} & 46.70 & 52.37 & 53.93 & 44.65 & 24.49 & 34.43 & \textbf{25.48} & 27.35 & 90.14 & 99.50 & 65.05 & 62.23 & 52.19 {\scriptsize\textcolor{red}{-0.6\%}} \\
\rowcolor{green!6}
  & + \papertitle\ (token) {\scriptsize(519)} & 46.59 & 52.25 & 54.04 & \textbf{46.03} & 25.62 & 34.75 & 25.31 & 27.08 & 90.60 & 99.50 & 64.95 & 61.76 & 52.37 {\scriptsize\textcolor{red}{-0.2\%}} \\
\rowcolor{green!6}
  & + \papertitle\ (block) {\scriptsize(1138)} & 46.10 & 52.65 & 55.39 & 43.83 & 27.86 & 33.73 & 25.01 & 27.24 & 91.47 & 99.50 & 65.00 & 62.22 & 52.50 {\scriptsize\textcolor{ForestGreen}{+0.0\%}} \\
\midrule
\rowcolor{blue!6}
Twilight & {\scriptsize Avg.\ 603} & 46.01 & 52.70 & 54.87 & 44.22 & 28.33 & 34.32 & 25.13 & 26.91 & 91.14 & 99.50 & 64.82 & \textbf{62.64} & 52.55 {\scriptsize\textcolor{ForestGreen}{+0.1\%}} \\
\rowcolor{green!6}
\papertitle (token) & {\scriptsize Avg.\ 636} & 45.61 & 52.78 & 54.01 & 45.59 & 28.40 & 34.09 & 25.11 & 27.12 & 91.05 & 99.50 & \textbf{65.31} & 62.25 & \textbf{52.57} {\scriptsize\textcolor{ForestGreen}{+0.2\%}} \\
\bottomrule
\end{tabular}%
}
\end{table*}

\begin{table*}[h]
\centering
\caption{Per-dataset accuracy of Qwen3-32B on LongBench (RTX 6000 Ada). \textbf{Bold} indicates the best result
per dataset. \textcolor{ForestGreen}{Green} and \textcolor{red}{red}
percentages denote change relative to Full Attention. The thinking mode is disabled for the experiment with \textit{enable\_thinking=False}. All top-$p$ configurations use $p$ = 0.95.}
\label{tab:longbench_ada_qwen3}
\renewcommand{\arraystretch}{1.1}
\resizebox{\textwidth}{!}{%
\begin{tabular}{@{}ll*{2}{c}*{3}{c}*{3}{c}cccc c@{}}
\toprule
& & \multicolumn{2}{c}{\textbf{Single-Doc.\ QA}} & \multicolumn{3}{c}{\textbf{Multi-Doc.\ QA}} & \multicolumn{3}{c}{\textbf{Summarization}} & \textbf{Few} & \textbf{Syn.} & \multicolumn{2}{c}{\textbf{Code}} & \\
\cmidrule(lr){3-4} \cmidrule(lr){5-7} \cmidrule(lr){8-10} \cmidrule(lr){11-11} \cmidrule(lr){12-12} \cmidrule(lr){13-14}
\textbf{Method} & \textbf{Budget} & Qasper & MF-en & HotpotQA & 2WikiMQA & Musique & GovRep. & QMSum & M-News & TriviaQA & PR-en & LCC & Repo. & \textbf{Avg.} \\
\midrule
\multicolumn{15}{c}{\cellcolor{gray!10}\textit{Qwen3-32B}} \\
\midrule
Full Attn. & 32k & 44.03 & 48.42 & 56.85 & 48.51 & \textbf{37.67} & 32.85 & 23.76 & 25.35 & \textbf{88.82} & \textbf{99.50} & \textbf{68.94} & 66.10 & 53.40 \\
\midrule
\multirow{5}{*}{Quest}
  & 256  & 27.56 & 34.51 & 25.50 & 24.34 & 11.50 & 24.70 & 22.18 & 24.13 & 65.84 & 77.00 & 53.80 & 43.25 & 36.19 {\scriptsize\textcolor{red}{-32.2\%}} \\
  & 1024  & 36.90 & 41.94 & 43.73 & 35.74 & 23.62 & 32.18 & 23.68 & \textbf{25.59} & 78.42 & 85.00 & 61.87 & 60.25 & 45.74 {\scriptsize\textcolor{red}{-14.3\%}} \\
  & 4096  & 42.24 & 47.05 & 54.81 & 45.90 & 33.86 & 32.56 & \textbf{24.20} & 25.19 & 87.86 & 87.00 & 68.58 & 65.72 & 51.25 {\scriptsize\textcolor{red}{-4.0\%}} \\
  & 8192  & 43.37 & 47.31 & 56.49 & 47.20 & 37.48 & 32.46 & 23.77 & 25.36 & 87.91 & 87.50 & 68.75 & 66.60 & 52.02 {\scriptsize\textcolor{red}{-2.6\%}} \\
\rowcolor{blue!6}
  & + Twi.\ {\scriptsize(429)} & 43.50 & 47.84 & 57.38 & 48.94 & 35.09 & 32.70 & 24.07 & 25.28 & 87.91 & 90.00 & 68.65 & \textbf{67.00} & 52.36 {\scriptsize\textcolor{red}{-1.9\%}} \\
\rowcolor{green!6}
  & + \papertitle\ (token) {\scriptsize(346)} & 43.01 & 47.77 & 57.42 & 49.37 & 35.92 & 32.54 & 23.64 & 25.26 & 87.74 & 89.00 & 68.62 & 66.20 & 52.21 {\scriptsize\textcolor{red}{-2.2\%}} \\
\rowcolor{green!6}
  & + \papertitle\ (block) {\scriptsize(1038)} & 43.12 & 48.59 & 57.14 & 48.81 & 36.38 & \textbf{33.09} & 23.77 & 25.16 & 88.74 & 99.00 & 68.77 & 66.52 & 53.26 {\scriptsize\textcolor{red}{-0.3\%}} \\
\midrule
\rowcolor{blue!6}
Twilight & {\scriptsize Avg.\ 535} & \textbf{44.67} & \textbf{48.78} & 57.04 & 49.13 & 36.39 & 32.86 & 23.73 & 25.31 & 88.74 & 99.00 & 68.59 & 65.84 & 53.34 {\scriptsize\textcolor{red}{-0.1\%}} \\
\rowcolor{green!6}
\papertitle (token) & {\scriptsize Avg.\ 449} & 44.52 & 48.47 & \textbf{58.05} & \textbf{49.87} & 36.03 & 33.08 & 23.37 & 25.30 & 88.74 & 99.00 & 68.74 & 65.74 & \textbf{53.41} {\scriptsize\textcolor{ForestGreen}{+0.0\%}} \\
\bottomrule
\end{tabular}%
}
\end{table*}

\begin{table}[htbp]
\centering
\caption{Comparison of KV Block Usage and Accuracy with and without Relocation for LongBench and LongBenchV2 (Llama-3.1-8B 128K). Results shown for both block sizes, 16 and 32. 'No Reloc.': the KV tokens keep the chronological order in the same way they were generated during prefill. 'Reloc': the KV tokens are relocated after the prefill phase. 'Reduction': Percentage decrease in the critical block count from 'No Reloc' to 'Reloc'.}
\label{tab:detailed_relocation_results_llama}
\scriptsize
\setlength{\tabcolsep}{3pt}
\resizebox{0.9\textwidth}{!}{%
\begin{tabular}{@{}lccccc@{\hspace{8pt}}ccccc@{}}
\toprule
 & \multicolumn{5}{c}{\textbf{Block Size: 16}} & \multicolumn{5}{c}{\textbf{Block Size: 32}} \\
\cmidrule(lr){2-6} \cmidrule(lr){7-11}
 & \multicolumn{2}{c}{\textbf{Block Count}} & \textbf{Reduction} & \multicolumn{2}{c}{\textbf{Accuracy}} & \multicolumn{2}{c}{\textbf{Block Count}} & \textbf{Reduction} & \multicolumn{2}{c}{\textbf{Accuracy}} \\
\cmidrule(lr){2-3} \cmidrule(lr){5-6} \cmidrule(lr){7-8} \cmidrule(lr){10-11}
\textbf{Category / Dataset} & \textbf{No Reloc.} & \textbf{Reloc.} & & \textbf{No Reloc.} & \textbf{Reloc.} & \textbf{No Reloc.} & \textbf{Reloc.} & & \textbf{No Reloc.} & \textbf{Reloc.} \\
\midrule
\multicolumn{11}{l}{\textit{LongBench (by Dataset)}} \\
narrativeqa            & 251.2 & 207.7 & 17.3\% & 28.05 & 27.93 & 146.4 & 116.6 & 20.4\% & 29.73 & 29.21 \\
qasper                 & 57.9  & 55.3  & 4.5\%  & 46.55 & 46.22 & 33.0  & 31.3  & 5.0\%  & 46.84 & 45.82 \\
multifieldqa\_en       & 68.0  & 63.1  & 7.2\%  & 52.53 & 53.10 & 39.2  & 36.0  & 8.4\%  & 52.01 & 51.71 \\
hotpotqa               & 116.4 & 102.4 & 12.0\% & 54.85 & 55.41 & 67.5  & 58.6  & 13.1\% & 51.74 & 52.07 \\
2wikimqa               & 68.3  & 62.9  & 7.9\%  & 46.52 & 45.06 & 39.2  & 35.8  & 8.8\%  & 45.73 & 45.29 \\
musique                & 141.1 & 123.1 & 12.7\% & 24.82 & 26.45 & 81.9  & 70.9  & 13.4\% & 22.81 & 21.93 \\
gov\_report            & 92.5  & 84.0  & 9.2\%  & 34.46 & 34.41 & 53.9  & 47.4  & 12.0\% & 34.00 & 33.56 \\
qmsum                  & 131.5 & 112.6 & 14.4\% & 25.35 & 25.27 & 77.3  & 63.7  & 17.7\% & 24.95 & 24.48 \\
multi\_news            & 32.8  & 32.3  & 1.7\%  & 27.43 & 27.38 & 19.1  & 18.7  & 1.9\%  & 27.45 & 27.52 \\
triviaqa               & 75.6  & 68.7  & 9.2\%  & 90.34 & 88.37 & 45.2  & 40.9  & 9.5\%  & 84.44 & 83.77 \\
passage\_retrieval\_en & 73.1  & 66.6  & 8.9\%  & 99.50 & 99.50 & 44.7  & 40.0  & 10.4\% & 99.50 & 99.00 \\
lcc                    & 38.0  & 36.9  & 2.9\%  & 63.32 & 63.35 & 22.1  & 21.4  & 3.3\%  & 62.98 & 63.09 \\
repobench-p            & 100.4 & 89.5  & 10.8\% & 59.67 & 58.67 & 58.8  & 51.1  & 13.0\% & 59.01 & 58.03 \\
\cmidrule(l){1-11}
\textbf{Average}       & \textbf{95.9} & \textbf{85.0} & \textbf{11.4\%} & \textbf{50.26} & \textbf{50.09} & \textbf{56.0} & \textbf{48.6} & \textbf{13.2\%} & \textbf{49.32} & \textbf{48.88} \\
\midrule
\multicolumn{11}{l}{\textit{LongBenchV2 (by Seq. Len.)}} \\
8K -- 16K              & 84.0  & 76.8  & 8.6\%  & 42.86 & 42.86 & 54.2  & 47.8  & 11.8\% & 39.29 & 42.86 \\
16K -- 32K             & 124.3 & 114.0 & 8.3\%  & 35.29 & 37.65 & 80.3  & 70.1  & 12.7\% & 29.41 & 28.24 \\
$>$ 32K                & 168.7 & 150.8 & 10.6\% & 14.29 & 14.29 & 110.8 & 98.9  & 10.8\% & 14.29 & 14.29 \\
\cmidrule(l){1-11}
\textbf{Average}       & \textbf{125.7} & \textbf{113.9} & \textbf{9.4\%} & \textbf{30.81} & \textbf{31.60} & \textbf{76.0} & \textbf{66.6} & \textbf{12.4\%} & \textbf{30.83} & \textbf{30.83} \\
\bottomrule
\end{tabular}}
\end{table}

\begin{table}[htbp]
\centering
\caption{Average accuracy on 11 RULER subtasks with Llama-3.1-8B-Instruct (50 samples per subtask). All configurations share the same prompts, tokenization, and scoring. \papertitle{} uses token-level top-$p$}
\label{tab:ruler}
\begin{tabular}{@{}lcc@{}}
\toprule
\textbf{Configuration} & \textbf{16K} & \textbf{32K} \\
\midrule
Full Attn. & 94.86 & 89.09 \\
Quest+Twilight & 94.88 & 88.66 \\
Twilight & 94.71 & 88.98 \\
Quest+\papertitle{} & 95.08 & 88.32 \\
\papertitle{} & 94.71 & 89.14 \\
\bottomrule
\end{tabular}
\end{table}

The 11 subtasks are the eight needle-in-a-haystack variants (single, multi-key, multi-value, and multi-query) plus variable tracking and common- and frequent-word extraction. On the eight needle subtasks, the \papertitle{} average is within 0.5 points of full attention at both lengths. Common-word extraction drops sharply at 32K for every configuration, including full attention (1.2\%): the model emits a generic stopword list regardless of the gold words, so this reflects a model limitation and not sparsity.

\begin{table}[htbp]
\centering
\caption{Comparison of KV Block Usage and Accuracy with and without Relocation for LongBench and LongBenchV2 (Qwen2.5-7B-Instruct, 32K). LongBench results are shown for block sizes 16 and 32. LongBenchV2 was evaluated at block size 16 only; $n$ is the number of samples per length bucket. 'No Reloc.': the KV tokens keep the chronological order in the same way they were generated during prefill. 'Reloc': the KV tokens are relocated after the prefill phase. 'Reduction': Percentage decrease in the critical block count from 'No Reloc' to 'Reloc'.}
\label{tab:detailed_relocation_results_qwen}
\scriptsize
\setlength{\tabcolsep}{3pt}
\resizebox{0.9\textwidth}{!}{%
\begin{tabular}{@{}lccccc@{\hspace{8pt}}ccccc@{}}
\toprule
 & \multicolumn{5}{c}{\textbf{Block Size: 16}} & \multicolumn{5}{c}{\textbf{Block Size: 32}} \\
\cmidrule(lr){2-6} \cmidrule(lr){7-11}
 & \multicolumn{2}{c}{\textbf{Block Count}} & \textbf{Reduction} & \multicolumn{2}{c}{\textbf{Accuracy}} & \multicolumn{2}{c}{\textbf{Block Count}} & \textbf{Reduction} & \multicolumn{2}{c}{\textbf{Accuracy}} \\
\cmidrule(lr){2-3} \cmidrule(lr){5-6} \cmidrule(lr){7-8} \cmidrule(lr){10-11}
\textbf{Category / Dataset} & \textbf{No Reloc.} & \textbf{Reloc.} & & \textbf{No Reloc.} & \textbf{Reloc.} & \textbf{No Reloc.} & \textbf{Reloc.} & & \textbf{No Reloc.} & \textbf{Reloc.} \\
\midrule
\multicolumn{11}{l}{\textit{LongBench (by Dataset)}} \\
narrativeqa            & 225.4 & 201.2 & 10.8\% & 19.83 & 16.52 & 125.9 & 113.3 & 10.0\% & 16.04 & 13.92 \\
qasper                 & 60.5  & 58.1  & 4.0\%  & 38.88 & 39.46 & 34.4  & 32.9  & 4.4\%  & 39.37 & 39.20 \\
multifieldqa\_en       & 71.4  & 67.2  & 5.9\%  & 48.38 & 49.62 & 40.6  & 37.7  & 7.1\%  & 48.79 & 48.98 \\
hotpotqa               & 118.0 & 107.7 & 8.7\%  & 51.89 & 51.02 & 66.8  & 60.9  & 8.9\%  & 49.99 & 48.54 \\
2wikimqa               & 71.7  & 67.3  & 6.2\%  & 34.03 & 34.55 & 41.0  & 38.1  & 7.0\%  & 35.05 & 34.52 \\
musique                & 140.4 & 128.8 & 8.3\%  & 22.66 & 21.75 & 79.2  & 72.2  & 8.8\%  & 21.74 & 18.14 \\
gov\_report            & 100.8 & 93.2  & 7.6\%  & 29.16 & 29.37 & 56.2  & 52.0  & 7.5\%  & 28.50 & 27.90 \\
qmsum                  & 121.2 & 107.7 & 11.2\% & 22.27 & 21.53 & 67.9  & 60.8  & 10.5\% & 21.95 & 22.28 \\
multi\_news            & 36.5  & 36.3  & 0.5\%  & 24.05 & 23.96 & 21.0  & 20.8  & 0.8\%  & 24.00 & 24.06 \\
triviaqa               & 86.3  & 79.9  & 7.4\%  & 82.06 & 81.21 & 49.6  & 46.8  & 5.8\%  & 80.42 & 78.71 \\
passage\_retrieval\_en & 67.5  & 63.8  & 5.5\%  & 91.88 & 91.92 & 39.1  & 37.2  & 4.9\%  & 92.03 & 91.54 \\
lcc                    & 38.0  & 37.1  & 2.2\%  & 47.34 & 48.39 & 21.9  & 21.4  & 2.2\%  & 47.37 & 47.79 \\
repobench-p            & 94.2  & 86.9  & 7.7\%  & 49.34 & 50.17 & 53.9  & 50.1  & 7.0\%  & 49.25 & 48.46 \\
\cmidrule(l){1-11}
\textbf{Average}       & \textbf{94.8} & \textbf{87.3} & \textbf{7.9\%} & \textbf{43.21} & \textbf{43.04} & \textbf{53.7} & \textbf{49.6} & \textbf{7.6\%} & \textbf{42.65} & \textbf{41.85} \\
\midrule
\multicolumn{11}{l}{\textit{LongBenchV2 (by Seq. Len.)}} \\
8K -- 16K (n=27)             & 90.9  & 84.4  & 7.1\%  & 33.33 & 33.33 & --    & --    & --     & -- & -- \\
16K -- 32K (n=89)            & 138.7 & 128.6 & 7.3\%  & 39.33 & 43.82 & --    & --    & --     & -- & -- \\
\cmidrule(l){1-11}
\textbf{Average}       & \textbf{114.8} & \textbf{106.5} & \textbf{7.2\%} & \textbf{36.33} & \textbf{38.58} & --    & --    & --     & -- & -- \\
\bottomrule
\end{tabular}}
\end{table}

\begin{table}[htbp]
\centering
\caption{Comparison of KV Block Usage and Accuracy with and without Relocation for LongBench (LongChat-7B-v1.5-32K, 32K). Block size 16. 'No Reloc.': the KV tokens keep the chronological order in the same way they were generated during prefill. 'Reloc': the KV tokens are relocated after the prefill phase. 'Reduction': Percentage decrease in the critical block count from 'No Reloc' to 'Reloc'.}
\label{tab:detailed_relocation_results_longchat}
\scriptsize
\setlength{\tabcolsep}{6pt}
\begin{tabular}{@{}lccccc@{}}
\toprule
 & \multicolumn{2}{c}{\textbf{Block Count}} & \textbf{Reduction} & \multicolumn{2}{c}{\textbf{Accuracy}} \\
\cmidrule(lr){2-3} \cmidrule(lr){5-6}
\textbf{Dataset} & \textbf{No Reloc.} & \textbf{Reloc.} & & \textbf{No Reloc.} & \textbf{Reloc.} \\
\midrule
narrativeqa            & 245.6 & 203.6 & 17.1\% & 19.90 & 18.48 \\
qasper                 & 68.4  & 63.9  &  6.5\% & 29.67 & 30.06 \\
multifieldqa\_en       & 85.5  & 76.6  & 10.4\% & 44.07 & 43.92 \\
hotpotqa               & 151.5 & 128.0 & 15.6\% & 32.78 & 33.10 \\
2wikimqa               & 94.1  & 83.3  & 11.5\% & 24.49 & 24.25 \\
musique                & 175.0 & 147.4 & 15.7\% & 13.98 & 13.60 \\
gov\_report            & 102.8 & 91.6  & 10.8\% & 31.63 & 31.07 \\
qmsum                  & 140.4 & 119.0 & 15.2\% & 22.65 & 22.96 \\
\cmidrule(l){1-6}
\textbf{Average}       & \textbf{132.9} & \textbf{114.2} & \textbf{14.1\%} & \textbf{27.40} & \textbf{27.18} \\
\bottomrule
\end{tabular}
\end{table}

\subsection{Additional Experiments}
\label{sec:additional_experiments}

\paragraph{Comparison with head-wise adaptive top-$k$.}
Table~\ref{tab:adakv} compares \papertitle{} with Ada-SnapKV, the strongest Ada-KV~\cite{feng2025adakv} baseline, on LLaMA-3.1-8B-Instruct. At a similar token budget, \papertitle{} achieves higher average accuracy. We leave comparison with Expected Attention and KVzip to future work.

\begin{table}[htbp]
\centering
\caption{Comparison with Ada-KV on 12 LongBench datasets (LLaMA-3.1-8B-Instruct). Numbers in the headers are token budgets.}
\label{tab:adakv}
\small
\begin{tabular}{@{}lcccc@{}}
\toprule
\textbf{Dataset} & \textbf{AdaKV-512} & \textbf{AdaKV-1024} & \textbf{Quest+\papertitle{} (520)} & \textbf{\papertitle{} (610)} \\
\midrule
Qasper   & 31.06 & 36.35 & 46.60 & 45.60 \\
MF-en    & 50.27 & 51.63 & 52.50 & 52.90 \\
HotpotQA & 53.90 & 54.97 & 54.80 & 55.10 \\
2WikiMQA & 43.48 & 43.95 & 46.50 & 44.70 \\
Musique  & 28.16 & 28.73 & 24.80 & 28.00 \\
GovRep.  & 22.02 & 24.28 & 34.50 & 33.90 \\
QMSum    & 23.28 & 24.16 & 25.40 & 25.20 \\
M-News   & 21.49 & 23.23 & 27.40 & 27.20 \\
TriviaQA & 91.67 & 91.48 & 90.30 & 90.60 \\
PR-en    & 99.50 & 100.00 & 99.50 & 99.50 \\
LCC      & 60.37 & 62.83 & 63.30 & 63.00 \\
Repo.    & 55.35 & 56.49 & 59.70 & 60.40 \\
\midrule
\textbf{Avg.} & 48.38 & 49.84 & 52.11 & 52.18 \\
\bottomrule
\end{tabular}
\end{table}

\paragraph{Colocated serving.}
We also adapted \papertitle{} to a colocated server and measured the maximum concurrent requests on LongBenchV2 with LLaMA-3.1-8B on one A100-40GB GPU, with 252 generated tokens on average (Table~\ref{tab:colocated}). The gain is smaller than in the disaggregated setting because the dense prefill phase affects colocated performance. 

\begin{table}[htbp]
\centering
\caption{Maximum concurrent requests in colocated serving.}
\label{tab:colocated}
\small
\begin{tabular}{@{}lccc@{}}
\toprule
\textbf{Prompt length} & \textbf{Dense} & \textbf{\papertitle{}} & \textbf{Ratio} \\
\midrule
8K  & 14 & 26 & 1.86$\times$ \\
16K & 8  & 14 & 1.75$\times$ \\
\bottomrule
\end{tabular}
\end{table}

\paragraph{CUDA Graph.}
vLLM captures a single static graph on one stream, which does not support the two-stream pipeline of \papertitle{}, so we disable graph capture in the main results. Multi-stream capture is supported by CUDA, and is a requested vLLM feature~\cite{vllm_issue_30211}. Disabling \texttt{torch.compile} and enabling two-stream CUDA Graph capture improves small-batch TPOT by up to 30\% on Llama-3.1-8B and Phi-3-mini. \texttt{torch.compile} must be disabled because TorchDynamo cannot treat a \texttt{torch.cuda.Stream} as a symbolic input.

%% file: appendix_limitations.tex
\section{Limitations and Future Work}
\label{sec:limitations}
\papertitle{} has several limitations that suggest avenues for future work.

\textbf{End-to-end performance at small batch sizes: } \papertitle{} is approximately 1.5$\times$ slower in TPOT than dense vLLM at batch sizes 1--4 (Figure~\ref{fig:combined_results}). Our two-stream pipeline is incompatible with vLLM's single-stream CUDA Graph implementation~\cite{vllm_issue_30211}, exposing per-kernel launch overhead, an engineering limitation rather than a fundamental one.
\textbf{INT4 selection fidelity: } The INT4 metadata cache introduces a small overshoot of 5--6 extra blocks at $p \in [0.8, 0.95]$ while maintaining $\geq 97\%$ recall (Table~\ref{table:int4_vs_fp16_top_p_comp}). The overshoot is benign for accuracy but slightly inflates bandwidth.
\textbf{Disaggregated-only evaluation:} We isolate the decode stage via disaggregated prefill/decode. Colocated results are in Appendix~\ref{sec:additional_experiments}; a full study of prefill interference is future work.
\textbf{One-shot token relocation:} The relocation kernel uses prefill attention scores as a static estimate of token importance. For long generation runs where attention patterns drift, periodic re-relocation may improve effectiveness. MoE serving and composition with KV quantization, prefix caching, and speculative decoding are left for future work.

%% file: appendix_B.tex
\clearpage
\section{Algorithms}
\label{label_appendix_B}

\input{algorithms/top_p_token}

\input{algorithms/top_p_full}

%% file: algorithms/top_p_token.tex
\begin{center}
\begin{minipage}{0.8\textwidth}
\hrule height 0.8pt
\vspace{4pt}
\refstepcounter{algorithm}
\noindent\textbf{Algorithm \thealgorithm} Token-Level INT4 Scoring + Top-$p$
\label{alg:token_topp}
\vspace{4pt}
\hrule height 0.4pt
\vspace{4pt}
\scriptsize
\linespread{0.9}\selectfont
\begin{algorithmic}[1]
\Procedure{TokenTopP}{$Q, K_{\text{INT4}}, P_{\text{tgt}}, B$}
    \Statex \textbf{Phase 1 (scoring kernel): Per-token scores} (context split across thread blocks)
    \State $m \gets -\infty$,\; $l \gets 0$
    \For{each tile $j$ of $K_{\text{INT4}}$-cache}
        \State $S_j \gets QK_j^\top / \sqrt{d}$ \Comment{INT4 dequant; mask tokens outside the selector}
        \State Store $S_j$ in the score buffer
        \State Update streaming $(m, l)$ with $S_j$
    \EndFor
    \State Store this split's partials $(m, l)$
    \Statex \textbf{Phase 2 (top-$p$ kernel): Histogram threshold} (one thread block per request and query head)
    \State $M, Z \gets$ merge partials $(m, l)$ across splits \Comment{global max and normalizer}
    \State $H \gets 0^{B}$
    \For{each token $t$}
        \State $p_t \gets \exp(S_t - M) / Z$ \Comment{on the fly, no separate softmax pass}
        \State $H[\mathrm{bucket}(p_t)] \mathrel{+}= p_t$ \Comment{bucket from the float bits of $p_t$}
    \EndFor
    \State $b \gets \max\{i : \mathrm{RevCumSum}(H)_i \ge P_{\text{tgt}}\}$
    \State $\tau \gets$ lower edge of bucket $b$
    \State \Return tokens $t$ with $p_t \ge \tau$
\EndProcedure
\end{algorithmic}
\vspace{4pt}
\hrule height 0.8pt
\end{minipage}
\end{center}

%% file: algorithms/top_p_full.tex
\begin{wrapfigure}{R}{0.55\textwidth}
\begin{minipage}{\linewidth} 
\hrule height 0.8pt
\vspace{4pt}
\refstepcounter{algorithm} 
\noindent\textbf{Algorithm \thealgorithm} Full Implementation of INT4 Top-$p$ Attention with Block Allocation
\label{alg:fused_attn_full} 
\vspace{4pt}
\hrule height 0.4pt
\vspace{4pt}
\scriptsize
\linespread{0.9}\selectfont 
\begin{algorithmic}[1]
\Procedure{TopPAttnAlloc}{$Q, K_{\text{int4}}, s, z, P_{\text{tgt}}, K_{\text{cap}}, B, W, \mathcal{T}_{\text{fetch}}$}
    \Statex \textbf{Phase 1: Block LSE computation \& running Top-K}
    \State $m \gets -\infty^{\text{BLOCK\_M}}$,\; $l \gets 0^{\text{BLOCK\_M}}$,\; $T_{\text{buf}} \gets \{-\infty\}^{K_{\text{cap}}}$
    \For{each tile $j \in [j_{\text{start}}, j_{\text{end}})$}
        \State Load $K_j^{\text{int4}}$, dequantize: $K_j \gets (\text{unpack}(K_j^{\text{int4}}) - z_j) \cdot s_j$
        \State $S_j \gets \text{scale} \cdot (Q_{\text{even}} K_j^{\text{even}\top} + Q_{\text{odd}} K_j^{\text{odd}\top})$
        \State Apply masks: softcap, causal, sliding-window, ALiBi, QQ-bias
        \State Reshape $S_j \in \mathbb{R}^{\text{BLOCK\_M} \times N_b \times \text{BLOCK\_SIZE}}$, $N_b = \text{TILE}/\text{BLOCK}$
        \State $m_j \gets \max_{\text{tile}} S_j$,\; $E_j \gets \sum \exp_2(S_j - m_j)$ \Comment{per-block sum}
        \State $V_j \gets m_j + \log_2(E_j)$ \Comment{block LSE, $\in \mathbb{R}^{\text{BLOCK\_M} \times N_b}$}
        \State Store $V_j \to$ \texttt{output\_attn\_ptr}
        \State $m_{\text{new}} \gets \max(m, m_j)$
        \State $l \gets l \cdot \exp_2(m - m_{\text{new}}) + \sum_b E_j \cdot \exp_2(m_j - m_{\text{new}})$
        \State $m \gets m_{\text{new}}$
        \State $T_{\text{buf}} \gets \text{TopK}_{K_{\text{cap}}}(\text{sort}(T_{\text{buf}} \mathbin\Vert V_j))$ \Comment{merge \& truncate}
    \EndFor
    \If{\texttt{decode\_id}[\text{seq}] $\ge \mathcal{T}_{\text{fetch}}$} \Return \Comment{skip allocation}\EndIf

    \Statex \textbf{Phase 2: Threshold $\tau$ via fast-path or LSE-domain histogram}
    \State $L_{\text{glb}} \gets m + \log_2(l)$
    \State $P_{\text{topK}} \gets \sum \exp_2(T_{\text{buf}} - L_{\text{glb}})$,\; $V_{\max} \gets \min(T_{\text{buf}}) - L_{\text{glb}}$
    \If{$P_{\text{topK}} \ge P_{\text{tgt}}$} \Comment{Fast path}
        \State $\tau \gets V_{\max} - \epsilon + L_{\text{glb}}$ 
    \Else \Comment{Slow path}
        \State $V_{\min} \gets V_{\max} - W$
        \For{each tile $j$ over all logical blocks}
            \State Reload $V_j$ from \texttt{output\_attn\_ptr};\; $\tilde{V}_j \gets V_j - L_{\text{glb}}$
            \State $p_j \gets \exp_2(\tilde{V}_j)$
            \State $P_{\text{above}} \mathrel{+}= \sum p_j \cdot \mathbb{1}[\tilde{V}_j > V_{\max}]$
            \State $b_j \gets \lfloor B \cdot \text{clip}((\tilde{V}_j - V_{\min})/W, 0, 1)\rfloor$
            \State $H[b_j] \mathrel{+}= p_j \cdot \mathbb{1}[\tilde{V}_j \in [V_{\min}, V_{\max}]]$
        \EndFor
        \State $C \gets \text{cumsum}(H)$;\; $R \gets \text{sum}(H) - C + H$ \Comment{reverse cumsum}
        \State $b^* \gets \max\{b : R[b] \ge P_{\text{tgt}} - P_{\text{above}}\}$ (or $-\infty$ if none)
        \State $\tau \gets V_{\min} + (b^*/B) \cdot W + L_{\text{glb}}$
    \EndIf

    \Statex \textbf{Phase 3: Block selection \& free-pool allocation}
    \For{each tile $j$ over all logical blocks}
        \State Reload $V_j$;\; $\text{heavy} \gets V_j \ge \tau$
        \State $\text{keep} \gets \text{heavy} \lor \text{is\_sink}(j) \lor \text{is\_local}(j)$
        \State $\text{need} \gets \bigvee_{\text{queries}} \text{keep}$ \Comment{reduce over BLOCK\_M}
        \State Load local \& remote block-table entries
        \State $\text{claim} \gets \text{need} \land (\text{local} < 0) \land (\text{remote} \ge 0)$
        \State $n \gets \sum \text{claim}$;\; $\text{base} \gets \text{atomicAdd}(\text{free\_pool\_head}, n)$
        \State $\text{idx} \gets \text{base} + \text{cumsum}(\text{claim}) - 1$
        \State $\text{new\_blk} \gets \text{free\_pool}[\text{idx}]$ where claimed
        \State Store $\text{new\_blk} \to$ \texttt{new\_block\_ids\_ptr}
    \EndFor
\EndProcedure
\end{algorithmic}
\vspace{4pt}
\hrule height 0.8pt
\end{minipage}
\vspace{-1.5em}
\end{wrapfigure}

%% file: checklist.tex
\section*{NeurIPS Paper Checklist}

\begin{enumerate}

\item {\bf Claims}
    \item[] Question: Do the main claims made in the abstract and introduction accurately reflect the paper's contributions and scope?
    \item[] Answer: \answerYes{} %
    \item[] Justification: All the claims made in the abstract and introduction are empirically verified in the evaluation section \ref{eval_section}. %

\item {\bf Limitations}
    \item[] Question: Does the paper discuss the limitations of the work performed by the authors?
    \item[] Answer: \answerYes{} %
    \item[] Justification: The paper discusses its potential limitations in Section~\ref{sec:limitations}.  %

\item {\bf Theory assumptions and proofs}
    \item[] Question: For each theoretical result, does the paper provide the full set of assumptions and a complete (and correct) proof?
        \item[] Answer: \answerNA{}
    \item[] Justification: This paper presents a systems contribution: an end-to-end integration of top-$p$ sparse attention into vLLM, including INT4 scoring and histogram top-$p$ kernels and a token relocation algorithm. The paper does not include formal theoretical results, theorems, or proofs. All claims are supported by empirical evaluation (Section~\ref{eval_section}).


    \item {\bf Experimental result reproducibility}
    \item[] Question: Does the paper fully disclose all the information needed to reproduce the main experimental results of the paper to the extent that it affects the main claims and/or conclusions of the paper (regardless of whether the code and data are provided or not)?
    \item[] Answer: \answerYes{}
    \item[] Justification: The paper provides detailed pseudocode for our contributed kernels: the INT4 scoring and histogram top-$p$ kernels (Algorithm~\ref{alg:fused_attn_short}, with the complete version in Appendix~\ref{label_appendix_B}) and the token relocation algorithm (Algorithm~\ref{alg:token_relocation}). Section~\ref{eval_section} specifies the exact model checkpoints (Llama-3.1-8B-Instruct, LLaMA2-7B-Chat, Longchat-7B-v1.5-32k, Qwen2.5-7B-Instruct, Qwen3-32B, Phi-3-mini-128k-instruct, all from public Hugging Face releases), benchmark datasets (LongBench~\cite{bai2024longbench}, LongBenchV2~\cite{bai2024longbench2}, RULER~\cite{hsieh2024ruler}, GSM8K~\cite{cobbe2021training}, COQA~\cite{reddy2019coqa}, PG-19~\cite{rae2019compressive}), hardware configuration, and hyperparameters including the top-$p$ threshold ($p=0.95$ for LLaMA-2/3.1, $p=0.85$ for LongChat), block size (16 tokens), $K_{\text{cap}}=8$, and relocation parameters (sorting distance 4096, sort regions 3). All baselines (Quest, DS, Twilight) are configured following their official repositories, with the same skip-first-two-layers convention applied for fair comparison.

\item {\bf Open access to data and code}
    \item[] Question: Does the paper provide open access to the data and code, with sufficient instructions to faithfully reproduce the main experimental results, as described in supplemental material?
        \item[] Answer: \answerYes{}
    \item[] Justification: The paper provides detailed algorithmic descriptions (Algorithm~\ref{alg:fused_attn_short}, Algorithm~\ref{alg:token_relocation}, and the full version in Appendix~\ref{label_appendix_B}) and complete experimental setup (Section~\ref{eval_section}), including model checkpoints (Llama-3.1-8B-Instruct, LLaMA2-7B-Chat, Longchat-7B-v1.5-32k, Qwen2.5-7B-Instruct, Qwen3-32B, Phi-3-mini-128k-instruct), benchmark datasets (LongBench, LongBenchV2, RULER, GSM8K, COQA, PG-19), hardware configuration, and hyperparameters. The code is available at \url{https://github.com/nnurlan008/DisSparse}.


\item {\bf Experimental setting/details}
    \item[] Question: Does the paper specify all the training and test details (e.g., data splits, hyperparameters, how they were chosen, type of optimizer) necessary to understand the results?
    \item[] Answer: \answerYes{} %
    \item[] Justification: All related details are in Section \ref{eval_section}. %

\item {\bf Experiment statistical significance}
    \item[] Question: Does the paper report error bars suitably and correctly defined or other appropriate information about the statistical significance of the experiments?
    \item[] Answer: \answerNo{} %
    \item[] Justification: We report accuracy from single runs and do not report error bars, due to the cost of repeating each configuration. Results can differ slightly between runs and GPU types. %

\item {\bf Experiments compute resources}
    \item[] Question: For each experiment, does the paper provide sufficient information on the computer resources (type of compute workers, memory, time of execution) needed to reproduce the experiments?
    \item[] Answer: \answerYes{} %
    \item[] Justification: Accuracy and kernel experiments ran on NVIDIA A100 and RTX 6000 Ada GPUs, PG-19 perplexity on an RTX PRO 6000 Blackwell GPU, and system experiments on two A100-40GB GPUs (Section~\ref{eval_section}). %
    \item[] Guidelines:
    \begin{itemize}
        \item The answer \answerNA{} means that the paper does not include experiments.
        \item The paper should indicate the type of compute workers CPU or GPU, internal cluster, or cloud provider, including relevant memory and storage.
        \item The paper should provide the amount of compute required for each of the individual experimental runs, as well as estimate the total compute. 
        \item The paper should disclose whether the full research project required more compute than the experiments reported in the paper (e.g., preliminary or failed experiments that didn't make it into the paper). 
    \end{itemize}
    
\item {\bf Code of ethics}
    \item[] Question: Does the research conducted in the paper conform, in every respect, with the NeurIPS Code of Ethics \url{https://neurips.cc/public/EthicsGuidelines}?
    \item[] Answer: \answerYes{} %
    \item[] Justification: The research conducted in the paper conforms with the NeurIPS Code of Ethics.%

\item {\bf Broader impacts}
    \item[] Question: Does the paper discuss both potential positive societal impacts and negative societal impacts of the work performed?
    \item[] Answer: \answerNA{} %
    \item[] Justification: This paper does not directly have societal impacts as it only optimizes model serving efficiency. However, by reducing the cost of LLM inference, it could indirectly amplify both the positive and negative societal impacts of LLMs deployed at scale.  %
    
\item {\bf Safeguards}
    \item[] Question: Does the paper describe safeguards that have been put in place for responsible release of data or models that have a high risk for misuse (e.g., pre-trained language models, image generators, or scraped datasets)?
    \item[] Answer: \answerNA{} %
    \item[] Justification: This paper does not release data or models that have a high risk for misuse. %

\item {\bf Licenses for existing assets}
    \item[] Question: Are the creators or original owners of assets (e.g., code, data, models), used in the paper, properly credited and are the license and terms of use explicitly mentioned and properly respected?
    \item[] Answer: \answerYes{} %
    \item[] Justification: The paper uses open-source repositories and publicly available models for the evaluation. (Section \ref{eval_section}). All the use of the existing assets in this paper respects the original licensing.

\item {\bf New assets}
    \item[] Question: Are new assets introduced in the paper well documented and is the documentation provided alongside the assets?
    \item[] Answer: \answerYes{} %
    \item[] Justification: We release our code at \url{https://github.com/nnurlan008/DisSparse}, with documentation in the repository. %

\item {\bf Crowdsourcing and research with human subjects}
    \item[] Question: For crowdsourcing experiments and research with human subjects, does the paper include the full text of instructions given to participants and screenshots, if applicable, as well as details about compensation (if any)? 
    \item[] Answer: \answerNA{} %
    \item[] Justification: This paper does not involve crowdsourcing nor research with human subjects. %

\item {\bf Institutional review board (IRB) approvals or equivalent for research with human subjects}
    \item[] Question: Does the paper describe potential risks incurred by study participants, whether such risks were disclosed to the subjects, and whether Institutional Review Board (IRB) approvals (or an equivalent approval/review based on the requirements of your country or institution) were obtained?
    \item[] Answer: \answerNA{} %
    \item[] Justification: This paper does not involve crowdsourcing nor research with human subjects. %

\item {\bf Declaration of LLM usage}
    \item[] Question: Does the paper describe the usage of LLMs if it is an important, original, or non-standard component of the core methods in this research? Note that if the LLM is used only for writing, editing, or formatting purposes and does \emph{not} impact the core methodology, scientific rigor, or originality of the research, declaration is not required.
    \item[] Answer: \answerNA{} %
    \item[] Justification: The core method and algorithm development of the paper does not involve LLM usage. %

\end{enumerate}